\documentclass{article}

\PassOptionsToPackage{numbers,compress}{natbib}
\usepackage[preprint]{neurips_2026}

\usepackage[utf8]{inputenc}
\usepackage[T1]{fontenc}
\usepackage{hyperref}[citecolor=lightblue]
\usepackage{url}
\usepackage{booktabs}
\usepackage{amsfonts}
\usepackage{amsmath}
\usepackage{amssymb}
\usepackage{nicefrac}
\usepackage{microtype}
\usepackage[table]{xcolor}
\arrayrulecolor{black!70}
\usepackage{graphicx}
\usepackage{enumitem}
\usepackage{subcaption}
\usepackage{multirow}
\usepackage{makecell}
\usepackage{tikz}
\usetikzlibrary{arrows.meta,positioning,shapes.geometric,fit,backgrounds}
\usepackage{pgfplots}
\pgfplotsset{compat=1.18}
\usepackage{pifont}    
\usepackage{bxcoloremoji}

\hypersetup{
   colorlinks = true,
    linkcolor = linkblue,
     urlcolor = linkblue,
    citecolor = mygreen,
    filecolor = blue,
   }
\renewcommand{\sectionautorefname}{Section}

\definecolor{mygreen}{RGB}{85,100,40}
\definecolor{softblue}{RGB}{90, 144, 180}
\definecolor{linkblue}{RGB}{41, 98, 155} 

\newcommand{\cmark}{\ding{51}}  
\newcommand{\xmark}{\ding{55}}  
\newcommand{\githublogo}{\raisebox{-1.5pt}{\includegraphics[height=1.05em]{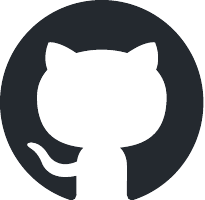}}}
\newcommand{\hflogo}{\raisebox{-1.5pt}{\includegraphics[height=1.05em]{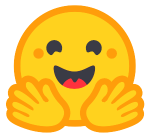}}}

\newcommand{\memgym}{\textsc{MemGym}}
\newcommand{\memgymdr}{\textsc{MemGym-DR}}
\newcommand{\memgymcodeqa}{\textsc{MemGym-CodeQA}}
\newcommand{\memrm}{\textsc{MemRM}}
\definecolor{lightblue}{rgb}{0.22,0.45,0.70}%
\title{\memgym{}: a Long-Horizon Memory Environment \\ for LLM Agents}

\author{%
  Wujiang Xu$^{1}$\thanks{Corresponding author: \texttt{wujiang.xu@rutgers.edu}}\quad Yu Wang$^{2}$\quad Kai Mei$^{1}$\quad Kaiqu Liang$^{3}$\quad
  Zhenting Wang$^{1}$\quad Mingyu Jin$^{1}$ \\[2pt]
  \bf Han Zhang$^{1}$\quad Shi-Xiong Zhang$^{2}$\quad Wenyue Hua$^{4}$\quad
  Sambit Sahu$^{2}$\quad Dimitris N. Metaxas$^{1}$ \\[6pt]
  \normalfont $^{1}$Rutgers University \quad $^{2}$Capital One \quad
  $^{3}$Princeton University \quad $^{4}$Microsoft Research
}

\begin{document}

\maketitle

\begin{abstract}
Memory is a central capability for LLM agents operating across long-horizon tasks. Existing memory benchmarks predominantly evaluate retention of personalized information in multi-turn chat scenarios, overlooking the dynamic memory formation that occurs during extended agent execution. Consequently, the memory systems they produce transfer poorly to realistic agentic environments such as coding and web navigation.
We present MemGym, a benchmark for agentic memory that unifies existing agent gyms and in-house memory-grounded pipelines behind one memory--reasoning interface. MemGym spans five evaluation tracks grouped into four agentic regimes: tool-use dialogue ($\tau^2$-bench), multi-turn deep-research search (\memgymdr{}), coding (SWE-Gym and \memgymcodeqa{}), and computer use (WebArena-Infinity). MemGym reports memory-isolated scores that decouple memory performance from reasoning, retrieval, and tool-use ability, so memory strategies can be ranked without those confounders.
Our synthetic pipelines for \memgymcodeqa{} and \memgymdr{} are length-controllable, ablation-verified at every stage, and tightly aligned with downstream scenarios.
To make evaluation on coding environments academically tractable, we train MemRM, a lightweight reward model (Qwen3-1.7B fine-tuned with QLoRA) that scores compression quality as a fast scalar read in place of full Docker rollouts.

\vspace{2mm}
{\centering
\coloremojicode{1F310}\,\href{https://wujiangxu.github.io/memgym-site/#}{\textbf{Project Page}} \quad
\githublogo\,\href{https://github.com/WujiangXu/MemGym}{\textbf{Code}} \quad
\hflogo\,\href{https://huggingface.co/MemGym}{\textbf{Dataset}}\par}
\end{abstract}

%


\begin{figure*}[tb!]
  \centering
  \includegraphics[width=0.9\linewidth]{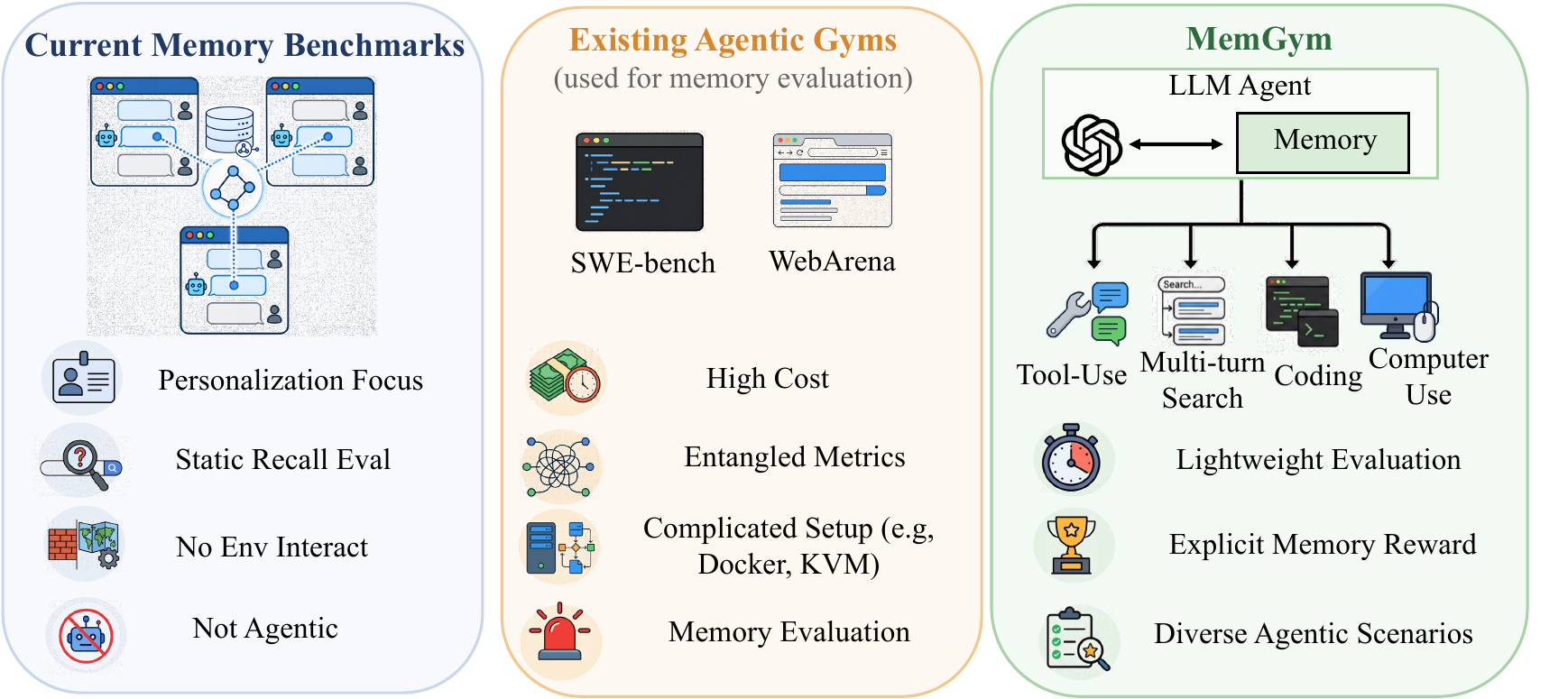}
  \caption{\memgym{} unifies five evaluation tracks across four agentic regimes (tool-use dialogue, multi-turn search, coding, computer use) behind a shared interface that separates memory from reasoning and supports memory-isolated scoring with explicit memory rewards.}
\label{fig:overview}
\end{figure*}
\section{Introduction}
\label{sec:intro}

LLM agents operating over long horizons must continuously decide what to preserve, summarize, or evict as observations, tool outputs, and intermediate conclusions accumulate. We refer to this process as \emph{memory formation during agent execution}, distinguishing it from the static recall tested by long-context benchmarks~\citep{bai2025longbench,hsieh2024ruler}. This capability arises across diverse realistic settings: coding agents revisit earlier debugging evidence across repository-scale tasks~\citep{jimenez2023swebench,pan2024swegym}, retrieval agents preserve bridge facts across search turns~\citep{trivedi2022musique,ho2020twowiki,yang2018hotpotqa}, and dialogue agents retain user constraints and tool state through extended interactions~\citep{yao2024taubench,barres2025tau2bench}.

Existing memory benchmarks~\citep{maharana2024locomo,wu2024longmemeval,hu2025memoryagentbench,ai2025memorybench} predominantly evaluate retention of personalized information in multi-turn chat, revealing little about memory behavior within agents that interleave perception, reasoning, and tool use. Three obstacles compound this gap. \textbf{(i)~Entangled metrics:} Agent gyms that involve long-horizon execution (SWE-Gym, $\tau^2$-bench, WebArena) report only end-task success, conflating memory failures with reasoning, retrieval, and tool-use errors. \textbf{(ii)~Illusory memory pressure:} Settings that appear memory-intensive often admit strong performance without explicit memory management, as facts remain re-derivable from repositories or recoverable from pretraining. \textbf{(iii)~Evaluation cost:} A single SWE-Gym rollout requires Docker infrastructure and tens of execution steps, placing systematic memory design iteration beyond most academic budgets.

We present \memgym{}, a benchmark, training-data pipeline, and lightweight evaluator that targets all three obstacles together (Figure~\ref{fig:overview}). \memgym{} unifies five evaluation tracks behind a shared interface that explicitly separates a memory module from the reasoning model: three wrappers around existing benchmarks ($\tau^2$-bench, SWE-Gym, WebArena-Infinity) plus two memory-grounded tracks we constructed in-house (\memgymdr{} deep research, \memgymcodeqa{} from SWE-smith). Every compression event is therefore observable, comparable, and recordable across scenarios. On top of this interface, \memgym{} reports \emph{memory-isolated} scores that disentangle memory performance from the underlying task, so that memory strategies can be ranked without being confounded by reasoning, retrieval, or tool-use ability.

Two complementary components close the loop from measurement to training. First, controllable synthetic pipelines for \memgymcodeqa{} and \memgymdr{} produce instances of tunable length and use verifier ablations to test intended memory against no-memory, distractor-only, and leakage-prone conditions; they are tightly aligned with the downstream coding and search scenarios they target, rather than acting as one-shot labor exercises. Second, \memrm{}, a lightweight reward model trained on collected trajectories (Qwen3-1.7B fine-tuned with QLoRA), scores compression quality as a fast scalar read in place of full Docker rollouts, which makes coding-environment evaluation academically tractable.
The same paired trajectories are released as a labeled corpus for downstream training research. \memgym{} therefore makes three contributions:

\begin{itemize}[leftmargin=1.5em]
  \item \textbf{Five tracks behind one memory interface, scored memory-isolated.} \memgym{} unifies $\tau^2$-bench, SWE-Gym, WebArena-Infinity, and the in-house \memgymdr{} and \memgymcodeqa{} pipelines under a shared memory contract, and reports paired baseline-vs-memory deltas under a fixed reasoner so the score reads as a memory effect rather than a confound of reasoning, retrieval, or tool use (\S\ref{sec:tracks}, \S\ref{sec:memrm}).
  \item \textbf{Controllable, ablation-verified synthetic pipelines.} \memgymcodeqa{} and \memgymdr{} generate length-tunable instances at scale, and verify via per-stage ablations that the intended memory channel (not parametric leakage or distractor recall) is the one being tested (\S\ref{sec:benchmark}).
  \item \textbf{\memrm{}: a scalar gate that replaces a Docker rollout.} A 1.7B-parameter QLoRA reward model trained on compression-event outcomes reaches AUROC $0.985$ on the SWE-Gym IID split, swapping a per-event rollout for a sub-second classifier call and supplying graded rewards for downstream post-training (\S\ref{sec:memrm}).
\end{itemize}

Together, these components turn the long-horizon evaluation loop from a one-way measurement into a closed feedback loop: the same trajectories that surface where current memory systems break also become the supervision signal for fixing them. The rest of the paper develops the framework (\S\ref{sec:method}), benchmark construction (\S\ref{sec:benchmark}), experiments (\S\ref{sec:experiments}), and future directions and limitations (\autoref{app:future-discussion-limitations}).

\begin{table}[t]
\centering
\caption{Comparison of memory and long-horizon agent benchmarks. \textit{Min.~Cost} is the cheapest path to score one memory configuration: Low, Medium, High.}
\label{tab:benchmark_comparison}
\setlength{\tabcolsep}{4pt}
\renewcommand{\arraystretch}{1.2}
\resizebox{\textwidth}{!}{%
\begin{tabular}{l l c c c c c c}
\toprule
\rowcolor{gray!20}
\textbf{Benchmark}
  & \textbf{Agentic Scenarios}
  & \shortstack{\textbf{\#}\\\textbf{Scn}}
  & \shortstack{\textbf{Inter-}\\\textbf{active}}
  & \shortstack{\textbf{Memory-}\\\textbf{Isolated}}
  & \shortstack{\textbf{Min.}\\\textbf{Cost}}
  & \shortstack{\textbf{Train.}\\\textbf{Data}}
  & \textbf{Length} \\
\midrule
\multicolumn{8}{l}{\textit{Dialogue-Centric Memory Benchmarks}} \\
LoCoMo~\cite{maharana2024locomo}               & Long-term dialogue       & 1 & \xmark & \xmark & Low    & \xmark & 9K          \\
LongMemEval~\cite{wu2024longmemeval}           & Long-term dialogue       & 1 & \xmark & \xmark & Medium & \xmark & 115K        \\
MemoryAgentBench~\cite{hu2025memoryagentbench} & Multi-turn dialogue      & 1 & \xmark & \xmark & Medium & \xmark & 100K--300K  \\
MemoryBench~\cite{ai2025memorybench}           & Continual dialogue       & 1 & \xmark & \xmark & Medium & \xmark & 30K--380K   \\
\midrule
\multicolumn{8}{l}{\textit{Long-Horizon Agent Benchmarks}} \\
SWE-Gym~\cite{pan2024swegym}                   & Repository coding          & 1 & \cmark & \xmark & High   & \cmark & Task-dep.     \\
$\tau^2$-bench~\cite{barres2025tau2bench}      & Tool-agent-user dialogue   & 1 & \cmark & \xmark & Medium & \xmark & Task-dep.     \\
WebArena-Infinity~\cite{webarena_infinity}     & Web computer use           & 1 & \cmark & \xmark & Medium & \xmark & Configurable  \\
\midrule
\multicolumn{8}{l}{\textit{Agent-Centric Memory Benchmarks}} \\
AMA-Bench~\cite{zhao2026amabench}              & Agentic apps (post-hoc QA) & 1 & \xmark & \xmark & Medium & \xmark & 57K           \\
AMemGym~\cite{amemgym2026}                     & Personalized conversation  & 1 & \cmark & \xmark & Medium & \xmark & Configurable  \\
\midrule
\textbf{\memgym{} (Ours)}
  & \shortstack[l]{Coding, web, tool-dialogue,\\deep-research search, coding QA}
  & \textbf{5} & \cmark & \cmark & \textbf{Low} & \cmark & \textbf{Configurable} \\
\bottomrule
\end{tabular}%
}
\vspace{-10pt}
\end{table}

\section{Related Work}
\label{sec:related}

\noindent\textbf{Agentic Memory Systems.}
Early memory-augmented LLM systems (MemoryBank~\citep{zhong2024memorybank}, MemGPT~\citep{packer2023memgpt}, and ReadAgent~\citep{lee2024readagent}) added explicit memory components but evaluated on long-term dialogue or document understanding rather than memory formation during environment interaction. A-Mem~\citep{xu2025amem} introduces agentic note evolution and is evaluated on LoCoMo~\citep{maharana2024locomo}, a long-term conversational-memory benchmark that mainly tests needle-in-a-haystack recall over personas and temporal event graphs rather than memory formed while debugging code, using tools, or navigating websites. LongMemEval~\citep{wu2024longmemeval}, MemoryAgentBench~\citep{hu2025memoryagentbench}, and MemoryBench~\citep{ai2025memorybench} extend this line with scalable histories, incremental multi-turn ingestion, and continual-learning feedback, but the memory target remains a transcript or feedback stream rather than a live trajectory under tool-use pressure. Most recently, AMA-Bench~\citep{zhao2026amabench} and AMemGym~\citep{amemgym2026} move closer to our setting (the former evaluates memory over agentic trajectories via post-hoc QA, the latter provides on-policy conversation with structured latent-state evolution), but neither offers a unified memory interface with memory-isolated rewards across coding, search, tool dialogue, and web control.

\noindent\textbf{Long-Horizon Agent Benchmarks.}
Long-horizon agent benchmarks evaluate whether agents can complete extended tasks in executable environments. SWE-bench~\citep{jimenez2023swebench} tests real GitHub issue resolution, and SWE-Gym~\citep{pan2024swegym} adds executable training tasks with unit tests and released trajectories; full evaluation is expensive and sparse, so end-task resolve rate alone is impractical for systematic memory iteration. $\tau$-bench and $\tau^2$-bench~\citep{yao2024taubench,barres2025tau2bench} evaluate tool-agent-user workflows but report end-task success without isolating whether failures came from memory, policy, or tool use. WebArena~\citep{zhou2023webarena} provides functional web environments, and WebArena-Infinity~\citep{webarena_infinity} scales this by automatically generating self-contained applications with verifiable tasks; OSWorld and OSGym~\citep{xie2024osworld,osgym2025} extend the same line to desktop and operating-system tasks. Across these benchmarks, memory is load-bearing but not separately measured. \memgym{} wraps such environments with an explicit memory boundary, records compression events, and reports memory-isolated scores so memory systems can be compared independently of the underlying agent's reasoning, retrieval, and tool-use ability.


\section{\memgym{}: A Memory-Centric Evaluation and Training Framework}
\label{sec:method}

\subsection{Overview}
\label{sec:overview}

\begin{figure}[t]
\centering
\includegraphics[width=0.9\linewidth]{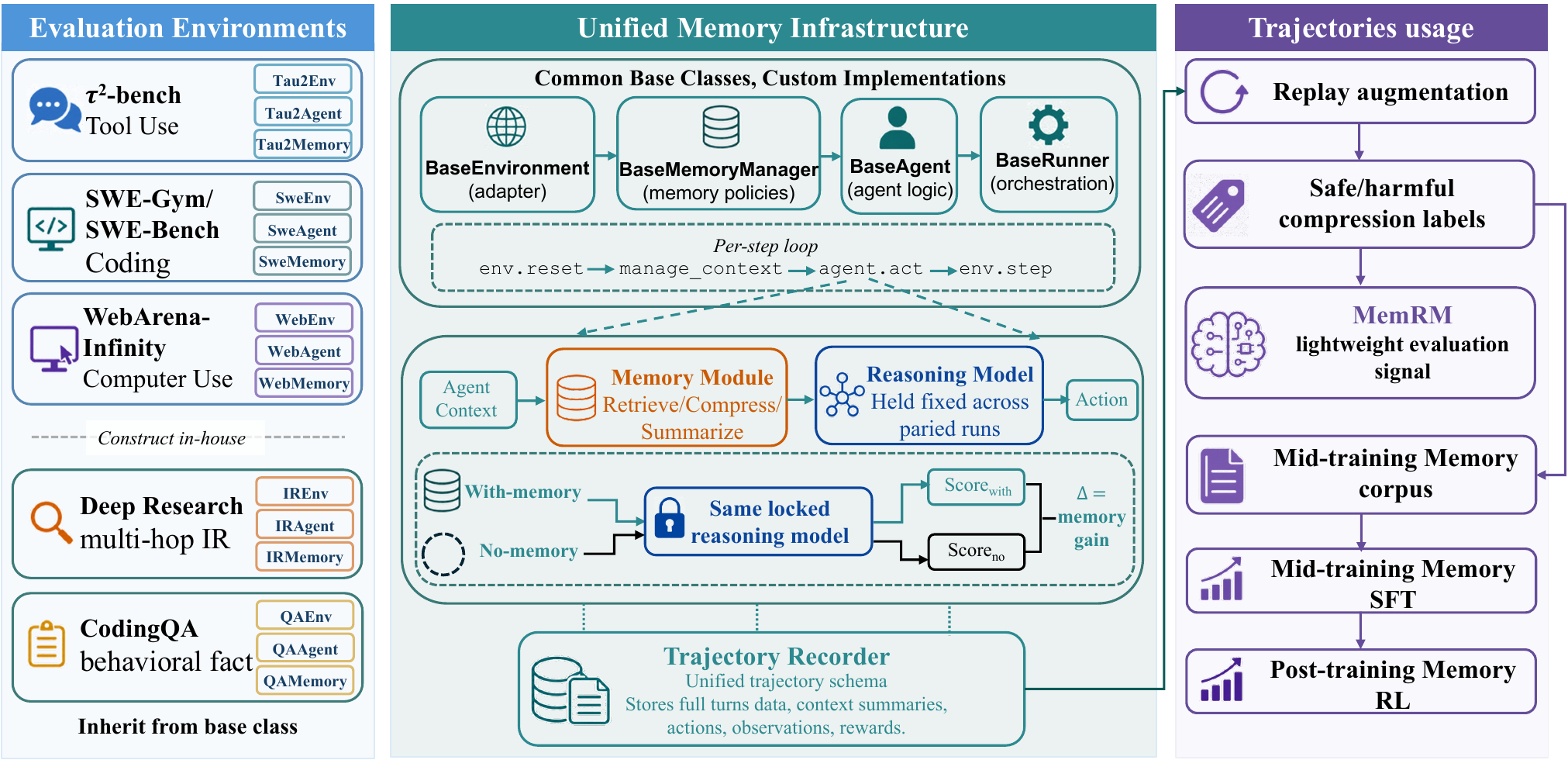}
\caption{\memgym{} architecture. Five environments share a memory module that wraps the prompt to the policy LLM, so the same strategy runs on any environment unchanged. Trajectories feed a replay-augmentation pipeline producing \textsc{safe}/\textsc{harmful} labels for \memrm{}; \memgymdr{} and \memgymcodeqa{} come from in-house pipelines (\autoref{sec:coding-qa-pipeline}).}
\label{fig:framework}
\vspace{-5pt}
\end{figure}

\memgym{} evaluates agentic memory across five environments unified by a shared memory module that wraps the prompt sent to the policy LLM: $\tau^2$-bench dialogue~\citep{barres2025tau2bench}, SWE-Gym coding~\citep{jimenez2023swebench,pan2024swegym}, WebArena-Infinity computer use~\citep{zhou2023webarena,webarena_infinity}, \memgymdr{} deep research, and \memgymcodeqa{}. The first three are wrappers around existing benchmarks; the latter two are environments for which we additionally constructed memory-grounded instances in-house, growing or extracting the facts the agent must retain rather than relying on benchmarks where memory state is incidental. All five plug into the same per-step contract described in \autoref{sec:tracks}; the two construction pipelines are themselves contributions and are detailed in \autoref{sec:coding-qa-pipeline}. Every with-memory vs.\ no-memory comparison in this paper holds the reasoning model fixed across both sides of the paired run, so the score delta (the \emph{memory gain}) isolates memory rather than confounding it with model choice. Trajectories collected through the unified wrapper feed a replay-augmentation pipeline that produces \textsc{safe}/\textsc{harmful} compression labels, which train \memrm{}, a lightweight classifier (Qwen3-1.7B fine-tuned with QLoRA) that predicts $\Pr[\text{behavior unchanged} \mid \text{compress}]$ in sub-second time, replacing per-episode Docker evaluation as the inner-loop signal for strategy iteration and serving as the reward signal for the post-training experiments in \autoref{sec:experiments}. Figure~\ref{fig:framework} shows how the pieces fit together.

\subsection{Unified Memory Infrastructure}
\label{sec:tracks}

Seven memory families plus a no-memory \emph{None} control are evaluated across $17$ (track, strategy) cells over the five environments (the cell list is enumerated in \autoref{app:track-strategy-grid}); all share one per-step contract, and the engineering depth (container backends, perturbation taxonomy, and harness patches) is documented in \autoref{app:harness-fixes}. All five environments plug into a common contract (\texttt{BaseMemoryEnvironment}, \texttt{BaseAgent}, \texttt{BaseMemoryManager}, \texttt{BaseRunner}) and a single per-step cycle: \texttt{env.reset()} $\to$ \texttt{memory\_manager.manage\_context} $\to$ \texttt{agent.act} $\to$ \texttt{env.step}. The memory manager wraps the prompt to the policy LLM and returns a \texttt{FilteredContext} plus a per-event \texttt{condensation\_event} (summary, forgotten-message indices, compression metadata); together with the per-step trajectory record (\autoref{app:gym-trajectory}), this reconstructs the training signal of any episode without re-running it. Per-environment wrappers capture each environment's dominant evaluation concern: $\tau^2$-bench wraps both the agent and the user simulator with independent memory managers, enabling the three-condition ablation that isolates whose compression matters; adding a new environment or strategy is a one-file change via \texttt{register\_env('name', cls)} or \texttt{register\_memory\_model('name', cls)}.

\label{sec:strategies}
Memory operations expose a single \texttt{manage\_context} contract, motivated by the OpenHands \texttt{CondenserPipeline}~\citep{wang2024openhands}, and compose via a \texttt{PipelineMemory} wrapper that accumulates per-stage statistics for downstream analysis. A \texttt{repair\_tool\_call\_pairs} primitive enforces the invariant that no compression can produce an API-invalid message sequence: if a summarizer drops the assistant turn that issued a tool call, the orphaned tool result is removed (and vice versa), since silently invalid sequences are silently rejected by Bedrock's Converse API and surface as zero-reward task failures rather than diagnosable errors. \memgym{} evaluates four memory-operation families as primary baselines (passthrough, LLM summarizing, structured per-environment summary, retrieval-style); the per-environment cross-product, one additional ported operation (observation masking), and the \texttt{set\_episode\_state} hook are documented in \autoref{app:strategies}.

\noindent\textbf{Replay-and-fork harness.}
A multi-runtime \texttt{ContainerBackend} covers Docker, Singularity (HPC), \texttt{BubblewrapBackend} (rootless namespaced sandbox over a Docker-exported rootfs, for clusters where neither Docker nor Singularity is available), \texttt{SwerexDockerBackend} (which copies recorded tool-result messages verbatim across the fork point because Bedrock's Converse API rejects orphaned \texttt{tool\_use} IDs), and a local \texttt{tempfile} backend; a \texttt{WebArenaServerPool} mirrors this on the WebArena side with port-range allocation, idle-pool reuse, and per-process log capture for parallel task execution. A \emph{replay-and-fork} protocol re-uses recorded tool actions to reconstruct repository state at the compaction step (\texttt{compute\_auto\_fork\_step} returns the earliest threshold-crossing step), and \texttt{ObservationReplayRunner}, which re-queries the policy LLM only on compaction-triggering steps, records a $10\times$ \texttt{policy\_call\_savings} that is what makes a 170-task hard-tier sweep tractable. The same recorded trajectories drive a counterfactual-replay augmentation pipeline realized as two processes from one rationale: cheap text-only replays from baseline trajectories for breadth, plus Docker-snapshotted replays from memory trajectories that yield ground-truth labels in milliseconds via \texttt{docker commit} rather than full re-execution. Together these expose the model at mid-training to what each memory operation causes \emph{in this scene}, so it learns per-step information importance rather than memorizing one canonical trajectory. Trajectory schemas, the harness fixes, and per-source label provenance are detailed in \autoref{app:gym-trajectory}, \autoref{app:harness-fixes}, and \autoref{app:memrm}; the score difference between paired baseline-vs-memory runs (always with the same reasoning model on both sides) is what we call \emph{memory gain} throughout the paper. Memory still alters the action distribution downstream of the wrapper, so memory gain attributes effects to the memory module under a fixed reasoner; it is not a clean ablation of memory as an independent capability.

\subsection{\memrm{} as a Lightweight Evaluation Signal}
\label{sec:memrm}

Measured on local trajectory data under Claude Sonnet~4.5, a single passthrough-memory episode costs \$2.10 on SWE-Gym~\citep{pan2024swegym} (median 55 turns and 700K cumulative input tokens), \$1.50 on WebArena-Infinity~\citep{zhou2023webarena} hard tier, \$0.40 on $\tau^2$-bench~\citep{barres2025tau2bench}, and \$0.15 on \memgymdr{} at list pricing of \$3/MTok input plus \$15/MTok output. A 100-episode sweep across the four interactive environments therefore costs \$420 and a 5-strategy~$\times$~3-seed sweep scales to \$6{,}300, making per-episode rollouts infeasible as the inner-loop signal for strategy iteration, ablation sweeps, or rollout filtering during training. We want a sub-second scalar predictor: given a (context-before, compressed-context, candidate-action) triple, estimate $\Pr[\text{behavior unchanged} \mid \text{compress}]$, usable as both a lightweight evaluation gate and the downstream reward signal in \autoref{sec:experiments}. It is a reward model in the RLHF sense~\citep{ouyang2022instructgpt}, not a latent-dynamics world model in the Ha--Schmidhuber sense~\citep{ha2018worldmodels}.

We obtain (context, action, label) training triples by \emph{replaying recorded trajectories at compaction events}, then aggregate \textsc{safe}/\textsc{harmful} labels from three complementary sources: (i)~episode-level task resolution on the parent trajectory, (ii)~counterfactual replay where a perturbation of the message window causes an action divergence on re-querying the policy, and (iii)~LLM-as-judge over the forgotten content. The pipeline runs two replay processes from one rationale: a cheap text-only process replays baseline trajectories to harvest counterfactual (context, action) pairs at scale, and a Docker-snapshotted process replays memory trajectories and verifies each perturbation's predicted action against a snapshot taken at the compaction event, producing ground-truth labels in milliseconds via \texttt{docker commit} rather than full re-execution. More details are in \autoref{app:memrm}.

\noindent\textbf{Training recipe.}
\label{sec:midtrain}
The augmented corpus contains $18.6$K (context, action, label) triples at a median context length of $22$K tokens; the train/eval split is grouped by repository (not by instance) to eliminate same-repo leakage. Each row carries the full agent view through the candidate compression and a binary completion (``\,Y\,'' for \textsc{safe}, ``\,N\,'' for \textsc{harmful}). \memrm{} is initialized from Qwen3-1.7B-Base and fine-tuned with QLoRA (NF4 4-bit, rank 16, $\alpha{=}32$, targets $\{q,k,v,o\}$) using TRL's \texttt{SFTTrainer} with \texttt{completion\_only\_loss=True}, so the loss fires only on the single label token. Class imbalance is handled by class-balanced cross-entropy with weights from the sklearn \emph{balanced} formula, capped at $w_{\max}=3.0$:
\begin{equation}
\label{eq:memrm}
\mathcal{L}_{\textsc{memrm}}(\theta) \;=\; -\,\mathbb{E}_{(c,y)\sim\mathcal{D}_\mathrm{aug}}\!\left[\,w_y \,\log p_\theta\!\bigl(y \mid c\bigr)\,\right], \qquad y \in \{\textsc{safe}, \textsc{harmful}\},
\end{equation}
where $c$ is the (context-before, compressed-context, candidate-action) triple linearized into the Qwen3 chat template and $w_y$ is the (capped) class-balanced weight. At inference time, we sweep a decision threshold $t^\star$ on held-out trajectories under the constraint $F_1^{\textsc{harmful}} \geq 0.90$ and $\mathrm{prec}^{\textsc{safe}} \geq 0.80$, and classify a candidate compression as \textsc{safe} iff $\Pr[\textsc{safe}\mid c] > t^\star$. \memrm{}'s primary use in this paper is as a lightweight evaluation gate (turning a 10-minute coding rollout into a sub-second scalar read) and as the post-training reward signal whose results we report in \autoref{sec:experiments}. Augmented-corpus statistics, the perturbation taxonomy, the env-feedback Docker snapshot protocol, the threshold sweep, and held-out calibration are in \autoref{app:memrm}.

\subsection{Constructed Pipelines for Memory-Grounded Evaluation}
\label{sec:coding-qa-pipeline}
\label{sec:benchmark}
\label{sec:memgymdr-pipeline}

\noindent\textbf{Template and sources.}
Two of the five evaluation environments (\memgymcodeqa{} and \memgymdr{}) additionally require constructing memory-grounded instances rather than wrapping an existing benchmark. Both pipelines follow the same template: ingest a source, extract a taxonomy of memory-only versus discoverable facts, inject shortcut-blocking distractors, scale to a target context length, and certify with a multi-criterion verifier. \memgymcodeqa{} ingests SWE-smith~\citep{yang2025swesmith} bug-and-patch instances; \memgymdr{} ingests results from academic search backends (arXiv, Semantic Scholar~\citep{kinney2023semanticscholar}, OpenAlex~\citep{priem2022openalex}, and Wikipedia).

\noindent\textbf{Fact extraction.}
What constitutes a ``memory-only fact'' differs by domain. \memgymcodeqa{} runs a three-pass extraction over each SWE-smith bug (gold-patch-visible seed, patch-hidden extraction, then discoverability re-examination with the full repository) and tags each fact as \texttt{discoverable} or \texttt{memory-only}; an instance is kept only if it exposes at least two critical \texttt{memory-only} facts (the threshold below which the question reduces to ``search the repo''). \memgymdr{} instead grows an ordered bridge-fact chain through iterative search, where each fact carries a \texttt{retention\_span} (the number of hops between when it is first discoverable and when it must be applied), so a 4-hop instance contains bridge facts requiring spans of $3$, $2$, and $1$, plus a terminal fact. The retention-span structure is the technical fingerprint that distinguishes multi-hop retention from single-pass retrieval.

\noindent\textbf{Distractors and length.}
Both pipelines inject shortcut-blocking distractors, but the shortcut taxonomy differs by domain. \memgymcodeqa{} blocks four shortcuts (topical inference, conventions-based inference, surface-pattern matching, and confidence-via-repetition) using cross-instance bug reports, same-repo docstrings, adversarial near-misses, and same-function contradictions; difficulty is then a composable post-hoc dial (prompt fuzzing, distractor scaling, indirection, fact fragmentation), so length and noise are decoupled from the underlying instance. \memgymdr{} injects a four-tier distractor hierarchy (natural / near-miss / adversarial-contradiction / bulk filler) and scales each instance to a target budget (10K--1M tokens). \memgymdr{} additionally requires a load-bearing \emph{fictionalization} stage: an LLM extracts entities (methods, models, organizations, people, numbers, years), generates fictional substitutes, and a deterministic regex applies the substitution registry uniformly across questions, answers, facts, documents, and distractors. Without fictionalization, frontier models score $0.70$--$0.85$ on \memgymdr{} by answering from pretraining alone; with it, no-memory drops to near-zero and the memory gap rises to $0.85$--$0.95$ (\autoref{app:ir-fictionalization}).

\noindent\textbf{Verifier and corpus.}
The verifier in \memgymcodeqa{} runs three independent checks per QA pair (solvability, distractor-confusion, and question-leakage), each catching a distinct shortcut; the three checks are jointly necessary because a single-check verifier had a $62\%$ false-positive rate. The verifier in \memgymdr{} runs a memory-ablation curve plus an adversarial-hack check, requiring \texttt{score\_all\_memory} $\geq$ \texttt{score\_long\_context} so that curated multi-hop notes must beat the long-context dump. The current corpora are 670 verified \memgymcodeqa{} instances (2{,}131 deduplicated QA pairs from a 1{,}000-instance candidate pool) and 1{,}194 verified \memgymdr{} instances (161 3-hop, 916 4-hop, 117 5/6-hop). The shared template (hidden gold artifact, memory-only/discoverable taxonomy, composable hardening, and multi-criterion verification) generalizes to any domain with a hidden ground-truth artifact and external documentation (medical guideline adherence, legal precedent application, scientific reproduction). Per-filter thresholds, the four-source \memgymcodeqa{} distractor taxonomy, the four-tier \memgymdr{} distractor hierarchy, the verifier pass criteria, and the SWE-Gym container backends are in \autoref{app:pipeline-details} and \autoref{app:ir-config}.

\section{Experiments: When Memory Does and Does Not Matter}
\label{sec:experiments}

The experiments map to three claims. (i)~Memory's payoff is regime-dependent across three wrapped gyms: roughly neutral on coding, where progress lives in the file system, and clearly positive on dialogue and web (\autoref{sec:wrapped-gyms}). (ii)~Under controlled pressure on two unrelated synthetic axes, the same strategy ranking reproduces and A-Mem leads at the maximum-pressure point on both (\autoref{sec:synthetic}). (iii)~\memrm{} ranks held-out compression events near-perfectly with calibrated probabilities and remains deployable on a characterized OOD subset (\autoref{sec:memrm-results}).

\subsection{Experimental Setup}
\label{sec:setup}

\textbf{Reasoners and gyms.} Wrapped-gym evaluations use Sonnet 4.5 on SWE-Gym (paired baseline-vs-memory via fork-batch replay over $1{,}003$--$1{,}041$ SWE-bench-style instances per reasoner), and Haiku 4.5 on $\tau^2$-bench (the $288$-task base split across mock, telecom, airline, and retail) and on WebArena-Infinity (a $140$-task hard slice across gmail, paypal, and gitlab; Playwright Chromium with text accessibility-tree observations and $\texttt{max\_steps}{=}50$). On SWE-Gym we additionally run Haiku 4.5 and GPT-OSS-120B as cross-reasoner controls. Trajectory compaction is triggered uniformly at $100$ messages or $32$K context tokens, whichever fires first, and the same trigger is used for every memory strategy in a comparison.

\textbf{Memory strategies and synthetic benchmarks.} We compare seven families: rolling Summary~\citep{wang2023scm}, A-Mem~\citep{xu2025amem} (note-evolution), MemoryBank~\citep{zhong2024memorybank}, LightMem~\citep{fang2025lightmem}, SimpleMem~\citep{liu2026simplemem}, Naive RAG~\citep{lewis2020retrieval} / BM25 retrieval (\memgymdr{} only), and a no-memory \emph{None} control. \memgymcodeqa{} stresses token budgets at $\{10\text{k}, 50\text{k}, 100\text{k}, 500\text{k}\}$ on the $4{,}289$-instance verified set built from SWE-smith repositories; \memgymdr{} stresses retrieval depth across $3$-, $4$-, and $5/6$-hop questions on a fictionalized $100$K-token deep-research pipeline. Both benchmarks use Sonnet 4.5 as the reasoner.

\textbf{\memrm{} training.} \memrm{} is Qwen3-1.7B-Base fine-tuned with QLoRA (NF4 4-bit, rank $16$, $\alpha{=}32$, target $\{q,k,v,o\}$ projections) on $18{,}642$ SWE-Gym compression-event labels at $32$K context, $600$ steps on $8\times$A100-40GB ($\sim$3 wall-clock hours). It is trained only on SWE-Gym compression events; OOD probes on memory-strategy and scenario axes (\autoref{tab:memrm}) are evaluation-only. Per-track configuration details (Playwright settings, $\tau^2$-bench base-split caveat, fictionalization mechanics, and the full \memrm{} recipe) are in \autoref{app:experimental-setup} and \autoref{app:memrm}.

\subsection{Memory Across Three Wrapped Gyms}
\label{sec:wrapped-gyms}

Across the three wrapped gyms, memory is roughly information-neutral on coding (where progress lives in the file system and the reasoner can re-read what was summarized away) and clearly beneficial on dialogue and web, where state hidden in past turns is expensive to re-derive (\autoref{tab:wrapped-gyms}).

\begin{table}[t]
\centering
\caption{Baseline and $+$memory resolve / success rates on three wrapped gyms (harness-verified). $\Delta$ in pp; \emph{Compr.}\ is the wrapper compression ratio averaged across compaction-triggering episodes.}
\label{tab:wrapped-gyms}
\small
\setlength{\tabcolsep}{5pt}
\begin{tabular}{lllrrrrr}
\toprule
\rowcolor{gray!20}
Gym & Model & Memory & $n$ & Baseline & $+$Memory & $\Delta$ & Compr. \\
\midrule
SWE-Gym           & Sonnet 4.5    & Summary    & 1041 & 42.8 & 42.8          & \phantom{$-$}0.0 & 1.47$\times$ \\
SWE-Gym           & Haiku 4.5     & Summary    & 1003 & 44.0 & 43.0          & $-$1.0           & 1.32$\times$ \\
SWE-Gym           & GPT-OSS-120B  & Summary    & 1003 & 22.3 & 19.1          & $-$3.2           & 1.45$\times$ \\
\midrule
$\tau^2$-bench    & Haiku 4.5     & Summary    &  288 & 50.0 & \textbf{58.7} & $+$8.7           & 2.29$\times$ \\
$\tau^2$-bench    & Haiku 4.5     & Structured &  288 & 57.6 & 60.1          & $+$2.5           & 1.86$\times$ \\
\midrule
WebArena-Infinity & Haiku 4.5     & Structured &  140 & 34.3 & \textbf{38.6} & $+$4.3           & 1.37$\times$ \\
WebArena-Infinity & Haiku 4.5     & Summary    &  140 & 34.3 & 35.0          & $+$0.7           & 1.45$\times$ \\
\bottomrule
\end{tabular}

\par\smallskip
{\footnotesize Per-track compression-ratio estimators, corpus composition, and pairing protocols are in \autoref{app:compr-estimators}.}
\vspace{-10pt}
\end{table}

On SWE-Gym, the resolve-rate change tracks reasoner strength rather than the memory mechanism: Sonnet 4.5 absorbs the lossy summary at zero cost ($\Delta{=}0$), Haiku 4.5 takes a small drop, and GPT-OSS-120B drops the most, reflecting weaker comprehension under compressed context. We do not see baseline-unsolvable instances rescued by adding memory: the coding tasks are difficult enough that the binding constraint is the reasoner, not the working-memory mechanism. What memory does buy on SWE-Gym is context compression (per-episode ratios of $1.32$--$1.47\times$ across the three reasoners), which keeps long trajectories within the policy's effective context without changing the resolve rate. On $\tau^2$-bench and WebArena-Infinity the picture inverts. Both involve state the reasoner cannot reliably hold across a long trajectory: multi-turn dialogue accumulates user constraints, open tool-call threads, and prior commitments hidden in earlier turns, while batch web operations require knowing which items have already been handled, information not in the rendered DOM. Once trajectories grow, memory summarizes the parts the reasoner can no longer keep in working context, and Haiku 4.5 answers correctly more often ($+8.7$ with Summary on $\tau^2$, $+4.3$ with Structured on WebArena). The effect size therefore tracks the cost of re-deriving discarded state: low for code, high for dialogue and web. The synthetic benchmarks below isolate that pressure on a single axis.

\subsection{Synthetic Memory Benchmarks: Head Ordering Under Pressure}
\label{sec:synthetic}

\begin{figure}[t]
\centering
\begin{minipage}[c]{0.5\linewidth}
\centering
\includegraphics[width=\linewidth]{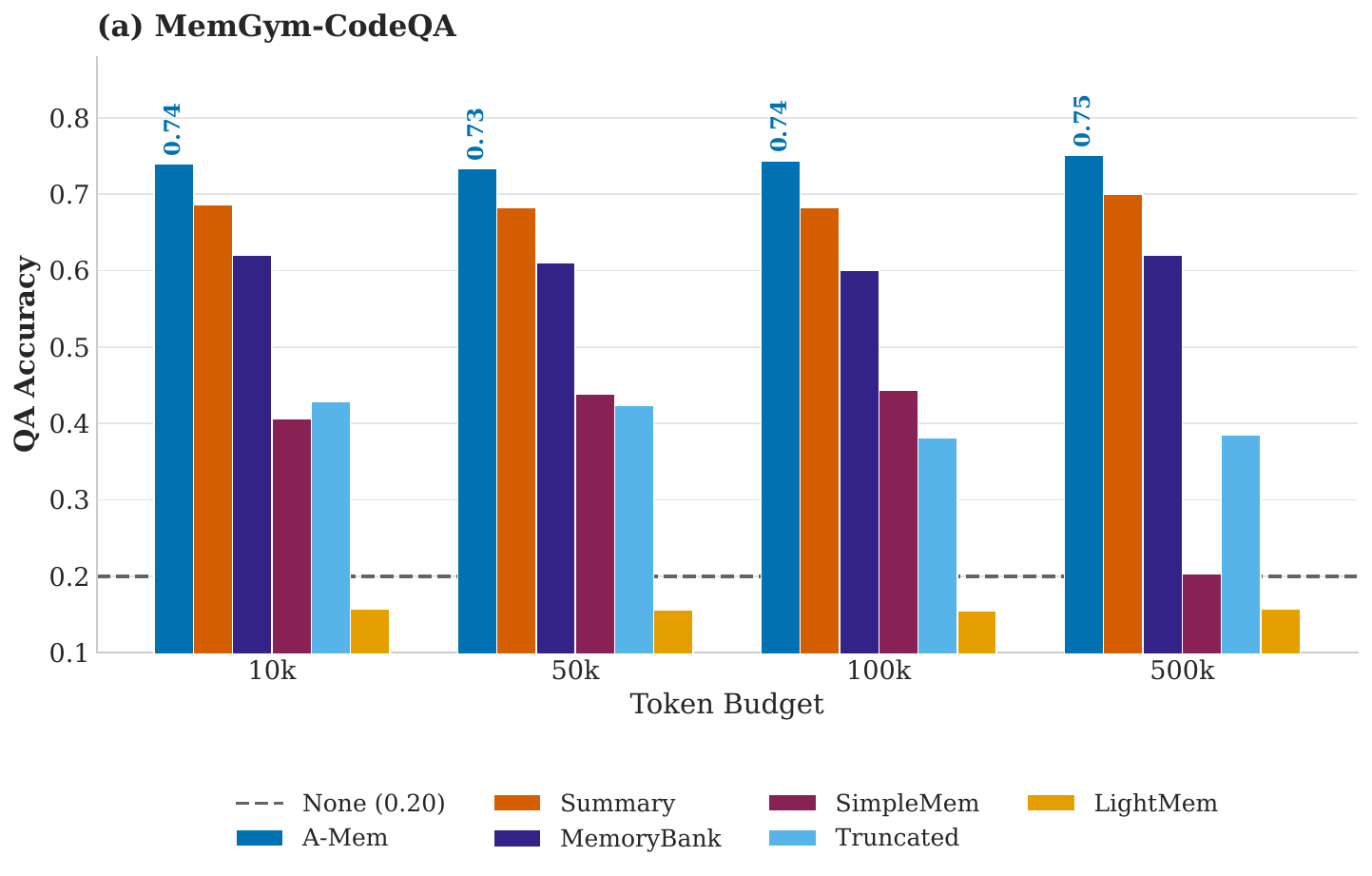}
\subcaption{\memgymcodeqa{} accuracy.}
\label{fig:synthetic-memory-a}
\end{minipage}\hfill
\begin{minipage}[c]{0.5\linewidth}
\centering
\footnotesize
\setlength{\tabcolsep}{4pt}
\renewcommand{\arraystretch}{1.15}
\begin{tabular}{lccc}
\toprule
\rowcolor{gray!20}
Strategy   & 3-hop & 4-hop & 5/6-hop \\
\midrule
A-Mem~\citep{xu2025amem}                 & 0.709 & 0.540 & \textbf{0.518} \\
BM25~\citep{robertson2009probabilistic}  & \textbf{0.808} & \textbf{0.555} & 0.425 \\
Naive RAG~\citep{lewis2020retrieval}     & 0.753 & 0.537 & 0.442 \\
MemoryBank~\citep{zhong2024memorybank}   & 0.699 & 0.537 & 0.482 \\
SimpleMem~\citep{liu2026simplemem}                               & 0.614 & 0.467 & 0.415 \\
LightMem~\citep{fang2025lightmem}        & 0.610 & 0.467 & 0.400 \\
None                                     & 0.330 & 0.290 & 0.009 \\
\bottomrule
\end{tabular}
\subcaption{\memgymdr{} judge score.}
\label{fig:synthetic-memory-b}
\end{minipage}
\vspace{-5pt}
\caption{Memory strategies on the two synthetic-memory benchmarks. \textbf{(a)} \memgymcodeqa{}: QA accuracy across token budgets ($10$k--$500$k); A-Mem bar values labelled and the no-memory \emph{None} baseline shown as a dashed reference. \textbf{(b)} \memgymdr{}: judge scores at $3$-, $4$-, and $5/6$-hop, with column-best in bold. A-Mem leads at the maximum-pressure point on both benchmarks ($500$k tokens on (a), $5/6$-hop on (b)); the strongest non-A-Mem baselines are rolling Summary~\citep{wang2023scm} on \memgymcodeqa{} and Naive RAG~\citep{lewis2020retrieval} / BM25~\citep{robertson2009probabilistic} on \memgymdr{}.}
\label{fig:synthetic-memory}
\vspace{-15pt}
\end{figure}

Wrapped gyms tell us \emph{when} memory pays off; the synthetic benchmarks ask which mechanism wins when the pressure is isolated to a single axis. \memgymcodeqa{} stresses the token budget from $10$k to $500$k on code; \memgymdr{} stresses retrieval depth from $3$ to $5/6$ hops on scientific text (\autoref{fig:synthetic-memory}). A-Mem is the best-performing strategy at the maximum-pressure point on both benchmarks, reaching $0.75$ on \memgymcodeqa{} at the $500$k-token budget ($+0.55$ vs.\ the no-memory baseline) and $0.518$ on \memgymdr{} at $5/6$-hop ($+0.509$ vs.\ baseline). The strongest non-A-Mem baselines are domain-dependent (rolling Summary on the coding QA axis and Naive RAG on retrieval), both of which A-Mem beats by a comfortable margin under maximum pressure.

\definecolor{mgAMEM}{HTML}{1F77B4}
\definecolor{mgRetA}{HTML}{2CA02C}
\definecolor{mgRetB}{HTML}{17A2A2}
\definecolor{mgTrunc}{HTML}{FF7F0E}
\definecolor{mgLight}{HTML}{D62728}
\definecolor{mgNone}{HTML}{7F7F7F}

\pgfplotsset{
  mgaxis/.style={
    width=\textwidth, height=5.0cm,
    tick label style={font=\scriptsize}, label style={font=\footnotesize},
    grid=both, grid style={gray!15}, major grid style={gray!22},
    every axis plot/.append style={line width=0.9pt, mark size=1.8pt},
    legend style={font=\scriptsize, draw=gray!40, fill=white, fill opacity=0.95,
                  text opacity=1, inner sep=2pt, row sep=-1pt,
                  at={(0.5,-0.32)}, anchor=north, legend columns=3, /tikz/every even column/.append style={column sep=6pt}},
    legend cell align=left,
  }
}

Without memory, both benchmarks collapse under pressure: the no-memory baseline reaches only $0.20$ on \memgymcodeqa{} and $0.009$ on \memgymdr{}'s $5/6$-hop slice (essentially random). Most flat baselines also lose ground as pressure rises; BM25 drops from $0.808$ at $3$-hop to $0.425$ at $5/6$-hop, confirming that the difficulty here is driven by the memory dimension rather than by task length per se. The payoff of memory therefore grows with pressure, and the gap is most visible on retrieval depth: A-Mem's lead over Naive RAG widens with hop count because note-evolution links bridge passages whose isolated query-relevance is low, which is the failure mode of a flat retriever at $5/6$-hop. On the coding axis the binding constraint is the strategy rather than the budget: every memory-equipped strategy is roughly flat across the four budgets ($10$k--$500$k), and LightMem's aggressive eviction policy actually underperforms the no-memory baseline on \memgymcodeqa{}, suggesting that policy quality matters more than window size for this working set. The same ordering reproduces on two unrelated domains with different bottlenecks; this consistency is the contribution rather than any single number. We make no domain-general claim on $n{=}2$ benchmarks, but the agreement makes A-Mem's lead unlikely to be a code-specific or scientific-text-specific artefact. Fictionalization mechanics and the no-fictionalization pilot are in \autoref{app:ir-fictionalization}.

\subsection{\memrm{}: A Learned Memory Critic}
\label{sec:memrm-results}

\memrm{} replaces a Docker rollout (minutes per event) with a sub-second classifier call that decides whether a candidate compression is \textsc{safe} to keep or \textsc{harmful} to revert. We report three standard metrics: AUROC (rank quality of the gate's \textsc{safe}-vs-\textsc{harmful} score, where $1.0$ is perfect and $0.5$ is chance), Expected Calibration Error (ECE; the gap between predicted and empirical \textsc{safe} rate over equal-mass probability bins), and Coverage (the fraction of out-of-distribution events that pass the per-axis selection rule). \autoref{tab:memrm} reports gate quality on a held-out SWE-Gym split.

\begin{table}[t]
\centering
\footnotesize
\setlength{\tabcolsep}{4pt}
\caption{\memrm{} gate quality on the SWE-Gym IID split and on two out-of-distribution axes (Qwen3-1.7B QLoRA). OOD rows report AUROC on the covered subset selected by a pre-declared per-axis selection rule; the rule, aggregate-OOD numbers, bootstrap protocol, and per-event detail metrics are in \autoref{app:memrm-ood} and \autoref{app:memrm-iid-detail}.}
\label{tab:memrm}
\rowcolors{2}{gray!8}{white}
\begin{tabular}{lrrrr}
\toprule
\rowcolor{gray!20}
Split & $n$ & \textbf{AUROC} [95\% CI] & Coverage & ECE \\
\midrule
SWE-Gym IID                                    & 3{,}007 & \textbf{0.985}                & --       & 0.009 \\
Strategy-OOD (sliding-window, masking, structured) & 166     & \textbf{0.714}~[0.54, 0.87]   & 26.5\%   & 0.850 \\
Scenario-OOD (WebArena V2)                     & 426     & \textbf{0.748}~[0.65, 0.86]   & 20.4\%   & 0.237 \\
\bottomrule
\end{tabular}
\vspace{-8pt}
\end{table}

\begin{figure}[t]
\centering
\includegraphics[width=\linewidth]{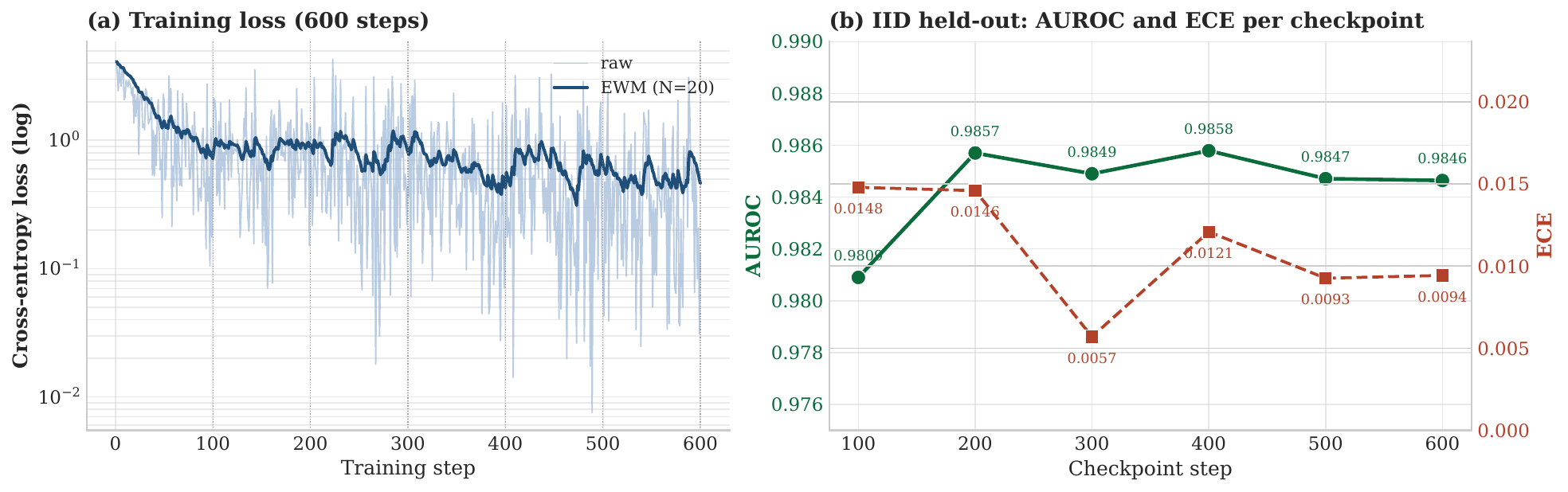}
\caption{\memrm{} training dynamics on SWE-Gym compression events (Qwen3-1.7B-Base + QLoRA, $32$K context, $600$ steps, $8\times$A100-40GB). }
\label{fig:rm-training-curves}
\vspace{-12pt}
\end{figure}

On the SWE-Gym held-out IID split, \memrm{} reaches AUROC $0.985$ with near-zero ECE: the gate ranks compression events near-perfectly and its predicted probabilities are calibrated. \autoref{fig:rm-training-curves} traces the training-time dynamics: cross-entropy descends steadily over $600$ steps and IID AUROC plateaus from step $200$ onward, so the headline metric is not the product of a lucky final checkpoint. We release this held-out split as part of the \memgym{} artifacts so other groups can evaluate their own gates against the same examples. \memrm{} also shows partial generalization beyond the training distribution along two axes. Within the coding domain, the gate transfers to memory strategies that did not appear in training (sliding-window, observation-masking, and structured memories), reaching AUROC $\approx 0.71$ at $\sim$$27\%$ coverage on the cohorts retained by a pre-declared per-axis selection rule based on selective classification~\citep{geifman2017selective}. Across domains, the gate transfers to WebArena V2 browsing trajectories and reaches AUROC $\approx 0.75$ at $\sim$$20\%$ coverage on a class-balanced covered subset. Aggregate AUROC over the full OOD sweeps is near-random and we do not claim deployment outside the covered subset; per-track \memrm{} variants are ongoing work, and the selection rule, polarity-flip diagnostics, and full training recipe are in \autoref{app:memrm}.
\section{Conclusion}
\label{sec:conclusion}

\memgym{} treats agentic memory as a first-class evaluation target rather than a number folded into task accuracy. Five tracks ($\tau^2$-bench, SWE-Gym, WebArena-Infinity, plus the in-house \memgymcodeqa{} and \memgymdr{} built from length-controllable, ablation-verified synthetic pipelines) plug into a common per-step contract that wraps the prompt before it reaches the policy LLM, so a memory module can be swapped without touching the reasoner; we score each run by the difference between paired baseline-vs-memory rollouts with the same reasoner on both sides. Alongside the wrappers, \memrm{} (Qwen3-1.7B fine-tuned with QLoRA) trades a multi-minute Docker rollout for a sub-second compression-quality call, which is what makes the full coding-environment evaluation academically tractable.
The picture our experiments paint is regime-dependent. On the wrapped gyms, memory's payoff tracks the cost of re-deriving discarded state: roughly neutral on coding (where progress lives in the file system) and clearly positive on dialogue and web, with mid- to high-single-digit-percentage success-rate gains. On the synthetic axes, the same strategy ranking reproduces on two unrelated domains at the largest pressure point on each, with A-Mem the most robust under maximum pressure where flat retrievers fail. \memrm{} achieves near-perfect IID ranking with calibrated probabilities on the SWE-Gym held-out split and remains deployable on a characterized covered subset of out-of-distribution traffic (\autoref{app:memrm-ood}).
We release the five-track wrappers, the \memgymcodeqa{} and \memgymdr{} synthetic pipelines, the \memrm{} weights, and the labeled paired-trajectory corpus as artifacts; downstream training recipes that consume these trajectories are future promising research directions.


\bibliographystyle{plainnat}
\bibliography{references}

\clearpage
\onecolumn
\tableofcontents

\clearpage
\appendix
\renewcommand{\sectionautorefname}{Appendix}

\section{Appendix Overview}
\label{app:overview}
The appendix is organized around the artifacts referenced in the main text. \autoref{app:strategies} documents the full memory-operation catalog including the operations $\times$ environments cross-product (\autoref{tab:strategies}). \autoref{app:gym-trajectory} gives the per-environment trajectory schemas, the multi-runtime container backends, the replay-and-fork protocol, and the observation-replay runner. \autoref{app:harness-fixes} lists the five SWE-Gym evaluation-harness bugs we patched. \autoref{app:pipeline-details} and \autoref{app:ir-config} together cover the two construction pipelines: per-filter rationales, three-pass behavioral fact extraction, and four-source distractor taxonomy for \memgymcodeqa{}; deep-research configuration, four-tier distractor hierarchy, and verifier pass criteria for \memgymdr{}. \autoref{app:memrm} gives the full \memrm{} training recipe (data, LoRA, threshold sweep, held-out metrics). \autoref{app:hyperparams} lists per-environment evaluation hyperparameters. \autoref{app:related-work} expands the main-text related-work section with detailed coverage of agentic memory systems, context compression, long-horizon agent benchmarks, and reward models. \autoref{app:experiments-detail} expands the main-text experiments with per-track setup, per-app and pilot breakdowns for the wrapped gyms, the \memgymdr{} fictionalization story plus the full per-strategy \memgymdr{} matrix, and the \memrm{} data-augmentation taxonomy.

\section{Extended Related Work}
\label{app:related-work}

\subsection{Agentic Memory Systems (Detailed)}
\label{app:related-memory}
The earliest line of memory-augmented LLMs treated memory as a way to compensate for finite context windows in long-form assistant settings. MemoryBank~\citep{zhong2024memorybank} stores and updates user memories for long-term companionship; MemGPT~\citep{packer2023memgpt} treats the prompt as a virtual memory hierarchy, paging between working and archival context; ReadAgent~\citep{lee2024readagent} builds episodic gist memories for long-document reading with optional look-up into the source text. Reflexion~\citep{shinn2023reflexion} introduced verbal self-reflection as an episodic-memory mechanism, in which the agent records natural-language critiques after each trial and conditions on them in subsequent attempts; this preceded the current ``memory-augmented agent'' line and remains a useful baseline mechanism for agent-side memory.

A broader line of work treats memory primarily as a \emph{prompting pattern} rather than an architectural change. Reflective and self-revision prompting (Reflexion~\citep{shinn2023reflexion}, Self-Refine~\citep{madaan2023selfrefine}, Self-Consistency~\citep{wang2023selfconsistency}, and Tree of Thoughts~\citep{yao2023tot}) uses prior outputs or critiques as in-context memory to guide later generations. Action-grounded reasoning prompts (ReAct~\citep{yao2023react} and Toolformer~\citep{schick2023toolformer}) interleave thought, action, and observation traces that themselves become a working-memory record. Persona- and role-driven memory has been studied through generative-agent simulations that combine episodic logs, reflective summaries, and retrieval-by-importance into a memory stream~\citep{park2023generative}, and through skill-library agents that distill repeated experience into reusable code or natural-language rules~\citep{wang2023voyager,zhao2024expel}. Hierarchical and graph-structured retrieval treats memory as an indexed knowledge structure rather than a flat history: tree-organized recursive abstraction~\citep{sarthi2024raptor}, community-summary graphs~\citep{edge2024graphrag}, and neurobiologically inspired hippocampal indexing~\citep{gutierrez2024hipporag}. Recursive summarization explicitly compresses dialogue history into a maintained summary state across turns~\citep{wang2023scm}. Production-oriented memory stacks~\citep{chhikara2025mem0} and parametric long-term memory architectures~\citep{wang2024memoryllm,das2024larimar} push memory either out to a managed service or into the model weights themselves. Retrieval-augmented self-critique extends memory use to the retrieval policy itself~\citep{asai2024selfrag}. These works span architecture, prompting strategy, and infrastructure; what they share is that the memory mechanism is benchmarked on a single domain (chat, code, web, or open-domain QA) using either end-task success or recall-style probes. \memgym{} is complementary: it fixes the memory interface across five tracks and uses paired memory-isolated scoring so any of the above mechanisms can be plugged in and compared on the same memory event.

A-Mem~\citep{xu2025amem} introduced an explicitly agentic memory mechanism: each new interaction is converted into a structured note, linked to related memories, and allowed to evolve existing memory representations. It evaluates on LoCoMo~\citep{maharana2024locomo}, a very-long-term conversational-memory benchmark built from 300-turn dialogues averaging 9K tokens across up to 35 sessions, with question answering, event summarization, and multimodal dialogue generation tasks. LongMemEval~\citep{wu2024longmemeval} adds six explicit memory competencies (information extraction, multi-session reasoning, knowledge update, temporal reasoning, single-session preference, abstention) over scalable conversational histories. MemoryAgentBench~\citep{hu2025memoryagentbench} adds incremental multi-turn ingestion to test how an agent updates memory as new information arrives. MemoryBench~\citep{ai2025memorybench} uses simulated user feedback to drive continual learning across sessions. REALTALK~\citep{lee2025realtalk} provides a 21-day real-world conversation dataset for long-term dialogue memory. MemoryArena~\citep{he2026memoryarena} evaluates agent memory in interdependent multi-session agentic tasks, where later sessions depend on facts established earlier. Most recently, AMA-Bench~\citep{zhao2026amabench} evaluates memory over real and synthetic agentic trajectories with QA, while AMemGym~\citep{amemgym2026} provides on-policy interactions with simulated users and structured latent-state evolution. Across these benchmarks, the memory target is a transcript, a feedback stream, or a personalized-dialogue state; none provides a unified memory interface with memory-isolated rewards across coding, search, tool dialogue, and web control.

\subsection{Context Compression and Long-Context Evaluation}
\label{app:related-compression}
A complementary line of work studies how to compress or evaluate long contexts in isolation, rather than how an agent should manage memory across an interaction. LLMLingua~\citep{jiang2023llmlingua} compresses prompts at the token level using a small language model to score perplexity, removing low-information tokens before the prompt is sent to the target model. AutoCompressors~\citep{chevalier2023autocompressors} learn soft summary vectors end-to-end so a long context can be replaced by a short sequence of compressed embeddings. Both target inference-time efficiency and assume a static prompt, in contrast to \memrm{}, which scores compression decisions taken during multi-step interaction.

Long-context evaluation is anchored by Needle-in-a-Haystack~\citep{kamradt2023needle}, RULER~\citep{hsieh2024ruler}, LongBench~\citep{bai2024longbench}, and LongBench v2~\citep{bai2025longbench}, which probe what an LLM can recall from a single long input. Multi-hop QA benchmarks (HotpotQA~\citep{yang2018hotpotqa}, 2WikiMultiHopQA~\citep{ho2020twowiki}, MuSiQue~\citep{trivedi2022musique}) and long-document QA such as QuALITY~\citep{pang2022quality} test compositional retrieval and reasoning over a fixed corpus. These benchmarks measure what can be \emph{recalled} from a static long context. Agentic memory must additionally decide \emph{what to preserve} across an interaction in which the context is generated turn-by-turn under tool-use pressure; \memgym{} measures the latter and uses memory-isolated paired rollouts to keep the comparison clean.

\subsection{Long-Horizon Agentic Benchmarks (Detailed)}
\label{app:related-benchmarks}
SWE-bench~\citep{jimenez2023swebench} evaluates agents on real GitHub issue resolution, with the original test split spanning thousands of bug reports across mature Python repositories. SWE-Gym~\citep{pan2024swegym} adds 2{,}438 Python training tasks with executable runtimes, unit tests, and released trajectories. We use SWE-Gym heavily because coding trajectories expose natural memory events: repeated evidence, stale observations, failed hypotheses, and patch-specific facts that survive multiple debugging steps. The cost is substantial: full evaluations require Dockerized repositories, test execution, and many interaction steps, so end-task resolve rate alone is too expensive and too sparse for systematic memory iteration. The OpenHands framework~\citep{wang2024openhands} provides the strategy-library lineage on which our memory-operation library is built.

$\tau$-bench and $\tau^2$-bench~\citep{yao2024taubench,barres2025tau2bench} evaluate tool-agent-user workflows in domains such as retail, airline, and telecom, but report end-task success without isolating whether failures came from memory, policy, or tool use. WebArena~\citep{zhou2023webarena} provides fully functional web environments across e-commerce, forums, collaborative software development, and content management; WebArena-Infinity~\citep{webarena_infinity} scales this idea by automatically generating self-contained web applications with verifiable tasks. These web benchmarks are realistic but require browser/server infrastructure and multi-step exploration, and their metrics conflate memory with navigation and planning. OSWorld~\citep{xie2024osworld} broadens computer-use evaluation to desktop and operating-system tasks under VM-style infrastructure with GUI grounding; OSGym~\citep{osgym2025} extends this to a gym-style training surface. Across these benchmarks, memory is load-bearing but not separately measured. \memgym{} wraps such environments with an explicit memory boundary, records compression events, and reports memory-isolated scores, enabling memory systems to be compared independently of the underlying agent's reasoning, retrieval, and tool-use ability.

\subsection{Reward Models and Learned Evaluators}
\label{app:related-reward}
Reward models trained from human or automated feedback have become a central tool for post-training LLMs~\citep{ouyang2022instructgpt}, and process reward models~\citep{lightman2023prm} extend this to step-level supervision over reasoning traces. \memrm{} is such a reward model specialized to memory: it supplies a step-level scalar that generalizes binary task success into a signal usable both for inference-time gating (a sub-second classifier replacing per-episode Docker rollouts) and for RL post-training as a dense reward channel. \memrm{} is deliberately not a world model in the Ha--Schmidhuber sense~\citep{ha2018worldmodels}, which predicts environment dynamics in a latent space; it predicts only whether a candidate compression would change behavior given the recorded action.

\section{Extended Experimental Details}
\label{app:experiments-detail}

This section expands \autoref{sec:setup} (per-track configuration), \autoref{sec:wrapped-gyms} (per-app and pilot rows), \autoref{sec:synthetic} (fictionalization mechanics for \memgymdr{}), and \autoref{sec:memrm-results} (\memrm{} data-augmentation taxonomy).

\subsection{Per-Track Experimental Setup}
\label{app:experimental-setup}

\noindent\textbf{Coding track.}
We evaluate on two SWE-Gym splits across 11 Python repositories: a 500-instance diverse subset spanning all repositories and a 554-instance long-trajectory subset drawn from issues whose baseline trajectory exceeds the compaction trigger. Most baseline numbers in \autoref{tab:wrapped-gyms} are reported on the combined $\sim$1{,}000-instance set; the per-strategy hyperparameter sweeps in \autoref{tab:hyperparam-ablation-full} use the smaller $n{=}20$ \texttt{moto} subset for cost reasons. The agent scaffold is mini-swe-agent v2.2.4 with a 250-step limit. Models: Claude Sonnet 4.5, Claude Haiku 4.5, and GPT-OSS-120B via Bedrock. For LLM-based memory strategies, the summarization model is Claude Haiku 4.5, reducing summarization cost by roughly $4\times$ compared to using the main reasoning model. Evaluation uses the official swebench harness, identical to the public leaderboard.

\noindent\textbf{\memgymdr{} track.}
We report the 100K-token deep-research pipeline with Claude Haiku 4.5 as the worker and Claude Sonnet 4.5 as the verifier and judge. Historical no-fictionalization runs diagnose parametric leakage; the Apr.~12 paper run characterizes the fictionalized pipeline at scale.

\noindent\textbf{Dialogue track.}
We evaluate on all 288 $\tau^2$-bench tasks across the mock, telecom, airline, and retail domains using Haiku 4.5 on Bedrock. Results are paired baseline-vs-memory runs in \texttt{mode=both}; the current headline numbers are on the base split, with official test-split reruns pending server-side support for \texttt{--task\_split\_name}.

\noindent\textbf{Web GUI track.}
We evaluate WebArena-Infinity hard tasks with Haiku 4.5, Playwright-controlled Chromium, text accessibility-tree observations, and a \texttt{max\_steps}=50 cap. The current web results use two protocols: smart replay on 170 hard tasks across five apps, and prefix injection on 140 hard tasks across three apps. Smart replay compares memory strategies after replaying recorded baseline actions; prefix injection fixes the early trajectory prefix and measures whether memory improves the live tail.

\noindent\textbf{Compute.}
\memrm{} training (Qwen3-1.7B QLoRA, full recipe in \autoref{app:memrm}) runs on $8\times$A100-40GB for $\sim$3~hours (${\approx}\,$2h45m training $+$ 22~min tokenization). Wrapped-gym evaluation is API-bound: per-task wall-clock is dominated by tool execution and reasoner latency rather than memory-manager compute. Preliminary and failed experiments not reported in the paper consumed roughly $2\times$ the reported training compute.

\subsection{Track-Strategy Evaluation Grid}
\label{app:track-strategy-grid}

\autoref{tab:track-strategy-grid} enumerates every (track, strategy) cell evaluated in the paper, with the summarizer and reasoner model used in each. Cells marked \checkmark{} were verified against the recorded run configuration; the only contaminated cell is \memgymdr{} $\times$ Structured (gpt-4o-mini summarizer used in error before Bedrock Haiku 4.5 became the documented default), reported in \autoref{tab:ir-detail} for completeness and excluded from the main-text figure.

\begin{table}[h]
\centering
\caption{Audit surface: (track, strategy) cells evaluated, with summarizer and reasoner. \checkmark{} = configuration verified; \texttimes{} = contaminated (one cell). \emph{None} is the no-memory control. Cells are dashes where the strategy is not applicable to the track (e.g., BM25/Naive RAG only apply to \memgymdr).}
\label{tab:track-strategy-grid}
\footnotesize
\setlength{\tabcolsep}{4pt}
\begin{tabular}{@{}lccccc@{}}
\toprule
Strategy & SWE-Gym & $\tau^2$-bench & WebArena-Inf. & \memgymcodeqa & \memgymdr \\
\midrule
None (control)        & \checkmark & \checkmark & \checkmark & \checkmark & \checkmark \\
Summary               & \checkmark & \checkmark & --         & \checkmark & --         \\
Structured            & \checkmark & \checkmark & \checkmark & --         & \texttimes \\
A-Mem                 & --         & --         & --         & \checkmark & \checkmark \\
MemoryBank            & --         & --         & --         & \checkmark & \checkmark \\
LightMem              & --         & --         & --         & \checkmark & \checkmark \\
SimpleMem             & --         & --         & --         & \checkmark & \checkmark \\
BM25 / Naive RAG      & --         & --         & --         & --         & \checkmark \\
Truncated             & --         & --         & --         & \checkmark & --         \\
\bottomrule
\end{tabular}
\end{table}

The grid contains $17$ verified cells and $1$ contaminated cell. Summarizer model is Bedrock Haiku 4.5 throughout (single exception above); reasoner is Sonnet 4.5 on SWE-Gym and \memgymcodeqa, Haiku 4.5 on $\tau^2$-bench and WebArena-Infinity, and Haiku 4.5 (worker) $+$ Sonnet 4.5 (verifier) on \memgymdr. Per-track hyperparameters are listed in \autoref{app:hyperparams}.

\subsection{Asset Licenses}
\label{app:asset-licenses}
Third-party assets are used under their respective open licenses. SWE-Gym~\citep{pan2024swegym}, SWE-bench~\citep{jimenez2023swebench}, $\tau^2$-bench~\citep{barres2025tau2bench}, SWE-smith~\citep{yang2025swesmith}, and the OpenHands condenser interface~\citep{wang2024openhands} are released under the MIT license. WebArena~\citep{zhou2023webarena}, the Qwen3-1.7B base model used for \memrm{}, and GPT-OSS-120B (used as one of the trajectory-collection reasoners) are released under Apache-2.0. Closed-source reasoners (Claude Sonnet~4.5 and Claude Haiku~4.5) are accessed via the AWS Bedrock API under Anthropic's commercial terms of use. The MemGym wrappers, the paired-trajectory corpus, and the synthetic \memgymcodeqa{} / \memgymdr{} instances are released under MIT; \memrm{} weights inherit the Apache-2.0 license of the Qwen3-1.7B base.

\subsection{Wrapped-Gym Compression-Ratio Estimators and Pairing Protocol}
\label{app:compr-estimators}

\noindent\textbf{Corpus composition and pairing.}
For SWE-Gym we pool the released diverse subset and the long-trajectory subset ($1{,}054$ instances across $11$ Python repositories). Instances whose baseline trajectory never reaches the compaction trigger ($\leq\!100$ messages) produce identical baseline and $+$memory rollouts and are therefore excluded from the paired comparison; the $+$Memory column in \autoref{tab:wrapped-gyms} reuses the baseline $n$ for these instances. The two $\tau^2$-bench baseline rows ($50.0$ vs.\ $57.6$ in \autoref{tab:wrapped-gyms}) are paired against independently sampled rollouts, so the within-row $\Delta$ is the paired statistic. WebArena-Infinity uses a $50\%$ prefix-injection ablation in which a randomly selected prefix of the trajectory is replayed verbatim into the agent before measurement begins.

\noindent\textbf{Per-track compression-ratio formulas.}
Let episode $e$ have $K_e$ compaction events and $r_{e,k}=T^{\text{pre}}_{e,k}/T^{\text{post}}_{e,k}$ denote the per-event input-token ratio (tokens before compaction divided by tokens after). Write $\mathcal{E}^{+}=\{e:K_e\geq 1\}$ for the set of compaction-triggering episodes and $T^{\max}_e$, $T^{\text{end}}_e$ for the peak and final token counts within episode $e$. The three reporters differ in which terms the wrapper persists:
\begin{align*}
\overline{r}_{\text{SWE}}
  &= \frac{1}{|\mathcal{E}^{+}|}\sum_{e\in\mathcal{E}^{+}}\frac{1}{K_e}\sum_{k=1}^{K_e} r_{e,k}, \\
\overline{r}_{\tau^2}
  &= \frac{1}{|\mathcal{E}^{+}|}\sum_{e\in\mathcal{E}^{+}} r_{e,K_e}, \\
\overline{r}_{\text{WA}}
  &= \frac{1}{|\mathcal{E}^{++}|}\sum_{e\in\mathcal{E}^{++}}\frac{T^{\max}_e}{T^{\text{end}}_e},
  \qquad \mathcal{E}^{++} = \bigl\{e\in\mathcal{E}^{+} : T^{\max}_e/T^{\text{end}}_e \geq 1.05\bigr\}.
\end{align*}
SWE-Gym averages every per-event ratio (wrapper field \texttt{memory\_stats.avg\_compression\_ratio}); $\tau^2$-bench reports only the last compaction event per episode (\texttt{wrapper\_last\_compression\_ratio}); WebArena derives the ratio from peak and final token counts at episode end and filters episodes whose agent terminated at peak (uninformative ratio of $1.0$). All three estimators are conditional on $K_e\geq 1$ so episodes that never triggered compaction do not pull the mean toward $1.0$.

\subsection{Wrapped Gyms: Per-App and Pilot Breakdowns}
\label{app:wrapped-gyms}

\noindent\textbf{From pairs to full coverage.}
The fork-batch protocol only forks instances whose baseline trajectory crosses the compaction-trigger threshold (more than $100$ messages). Below that threshold no compaction event fires and the $+$memory rollout is bit-identical to the baseline, so only a fraction of the full $\sim\!1{,}000$-instance evaluation set produces a non-trivial paired comparison: $678/1041$ ($65.1\%$) for Sonnet 4.5, $653/1003$ ($65.1\%$) for Haiku 4.5, and $556/1003$ ($55.4\%$) for GPT-OSS-120B. The lower coverage on GPT-OSS reflects its higher rate of early termination at the $250$-step ceiling rather than a memory effect. Reporting the $+$Memory rate against the full evaluation set requires extrapolating the paired delta back over the entire scale baseline:
\begin{equation*}
\begin{aligned}
\text{full}\ +\!\text{memory resolved}
  \;=\;& \text{paired memory resolved} \\
       & {}+ \bigl(\text{scale baseline resolved} - \text{paired baseline resolved}\bigr),
\end{aligned}
\end{equation*}
which is equivalent to the weighted mean $\Delta_{\text{full}} = \Delta_{\text{paired}} \times (n_{\text{paired}} / n_{\text{scale}})$. Plugging in the values above yields $446/1041$, $431/1003$, and $192/1003$ as the full-1k $+$memory resolution counts for Sonnet 4.5, Haiku 4.5, and GPT-OSS-120B respectively. All conditional $\Delta$ statistics (R$\to$R, R$\to$U, U$\to$R, U$\to$U transition counts; per-split breakdowns) are reported on the paired subset only.

\noindent\textbf{Hyperparameter sweep.}
\autoref{tab:hyperparam-ablation-full} reports a Sonnet 4.5 sweep over five strategy families (free-text summarization, selective observation masking, sliding-window, structured summarization at three trigger densities, and a chained masking-then-structured pipeline) against a fixed $n{=}20$ \texttt{moto} subset whose baseline resolves $12$ of $20$ instances ($60\%$). The summarizer model is held at Haiku 4.5 throughout. None of the strategies improves resolve rate at the pilot size, but they differ substantially in compression ratio achieved and in how often they trip the limits-exceeded ceiling, providing a controlled view of the cost side of memory.

\begin{table}[t]
\centering
\caption{Extended memory configuration sweep on SWE-Gym (Sonnet 4.5, \texttt{moto} subset, $n{=}20$, baseline $12/20 = 60\%$, Haiku 4.5 summarizer). \emph{Resolved}: harness verdict on the final patch; \emph{LimExc}: trajectories hitting the 250-step ceiling. The two columns are independent and a trajectory may count under both. Run IDs map to \texttt{experiments.md}.}
\label{tab:hyperparam-ablation-full}
\footnotesize
\setlength{\tabcolsep}{4pt}
\rowcolors{2}{gray!8}{white}
\begin{tabular}{@{}llp{4.2cm}rrrr@{}}
\toprule
\rowcolor{gray!20}
Run & Strategy & Params & Resolved & LimExc & Compress & $\Delta$ vs base \\
\midrule
6a & none (baseline)             & ---                          & 12 (60\%) &  0 & 1.00$\times$ & \phantom{$-$}0\phantom{0} pp \\
6c & structured\_summary         & ms=100, r=0.75, kf=3         & 12 (60\%) &  5 & 1.50$\times$ & \phantom{$-$}0\phantom{0} pp \\
6d & structured\_summary         & ms=200, r=0.75, kf=1         & 12 (60\%) &  0 & 1.00$\times$ & \phantom{$-$}0\phantom{0} pp \\
6e & structured\_summary         & ms=100, r=0.5, kf=1          & 11 (55\%) & 14 & 2.28$\times$ & $-5$\phantom{0} pp \\
6b & observation\_masking        & w=100                        & 10 (50\%) &  7 & 1.16$\times$ & $-10$ pp \\
7a & llm\_summarizing            & ms=100, r=0.75, kf=3         & 12 (60\%) &  0 & 1.35$\times$ & \phantom{$-$}0\phantom{0} pp \\
7b & observation\_masking (sel.) & w=100, kf=3                  & 12 (60\%) &  7 & 1.08$\times$ & \phantom{$-$}0\phantom{0} pp \\
7c & sliding\_window             & w=75, kf=3                   & 12 (60\%) &  9 & 1.23$\times$ & \phantom{$-$}0\phantom{0} pp \\
7e & pipeline (mask$\to$struct)  & w=100, ms=100, r=0.75, kf=3  & 12 (60\%) &  4 & 1.42$\times$ & \phantom{$-$}0\phantom{0} pp \\
\bottomrule
\end{tabular}
\end{table}

\noindent\textbf{$\tau^2$-bench gains are domain-conditional.}
The aggregate $+8.7$pp Summary gain on $\tau^2$-bench masks substantial domain heterogeneity. \autoref{tab:tau2-detail} disaggregates the gain across the four base-split domains (mock, telecom, airline, retail; $n{=}288$): Summary memory contributes its largest paired delta on telecom ($+17.5$pp on $n{=}114$) and a near-zero delta on retail, while structured memory wins on airline ($+14.0$pp on $n{=}50$). The per-domain $\Delta$ in each row uses that strategy's own independently sampled baseline rollouts (consistent with the cross-row caveat noted under \autoref{tab:wrapped-gyms}). Compression columns report the per-episode last-event ratio averaged over compaction-triggering episodes within each domain; the Total cells reproduce the run-level aggregator from \autoref{tab:wrapped-gyms} and may differ slightly from a domain-weighted mean since the runtime accumulates the global last ratio at run scope rather than averaging per-domain task means.

\begin{table}[t]
\centering
\caption{$\tau^2$-bench per-domain results with Haiku 4.5 (base-split, $n{=}288$ over mock/telecom/airline/retail). Baseline columns pair with the corresponding strategy row's independently sampled baseline rollouts (notes of \autoref{tab:wrapped-gyms}); $\Delta$ is the within-row paired statistic. Last-ratio columns are the episode-mean of \texttt{wrapper\_last\_compression\_ratio} over compaction-triggering episodes; mock has zero compaction events under either strategy at this $kl$ horizon (dialogues are short). Total ratios are reproduced from \autoref{tab:wrapped-gyms}'s run-level aggregator. Numbers from \texttt{probe\_tau2\_per\_domain.py}.}
\label{tab:tau2-detail}
\footnotesize
\setlength{\tabcolsep}{4pt}
\begin{tabular}{lrrrrrrr}
\toprule
\rowcolor{gray!20}
\multicolumn{1}{l}{} & \multicolumn{1}{c}{} & \multicolumn{2}{c}{Summary (paired)} & \multicolumn{2}{c}{Structured (paired)} & \multicolumn{2}{c}{Compr. (last-event)} \\
\cmidrule(lr){3-4}\cmidrule(lr){5-6}\cmidrule(lr){7-8}
\rowcolor{gray!20}
Domain & $n$ & $+$Summ & $\Delta$ & $+$Struct & $\Delta$ & Summ & Struct \\
\midrule
mock     &  10 & 50.0 & \phantom{$+$}0.0\phantom{0} & 50.0 & \phantom{$+$}0.0\phantom{0} & --- & --- \\
telecom  & 114 & 57.0 & $+17.5$           & 50.0 & $+$1.8\phantom{0}            & 1.38$\times$ & 1.24$\times$ \\
airline  &  50 & 54.0 & $+$6.0\phantom{0}            & 70.0 & $+14.0$           & 2.45$\times$ & 1.76$\times$ \\
retail   & 114 & 63.2 & $+$1.8\phantom{0}            & 66.7 & $-$1.8\phantom{0}            & 2.46$\times$ & 2.05$\times$ \\
\midrule
\textbf{Total} & \textbf{288} & \textbf{58.7} & \textbf{$+$8.7} & \textbf{60.1} & \textbf{$+$2.5} & \textbf{2.29$\times$} & \textbf{1.86$\times$} \\
\bottomrule
\end{tabular}
\end{table}

The mock domain is an LLM-simulated user (rather than the canonical scripted user) and has historically been the highest-variance slice of $\tau^2$; readers comparing to scripted-domain numbers should weight it accordingly. The dominant Summary gain ($+17.5$pp) sits on telecom, where dialogues are long enough to compact 18 times per episode (median) but the per-episode last-ratio (1.38$\times$) is among the least aggressive, consistent with kept-floor logic limiting how much can be pruned. Airline shows the largest Structured gain ($+14.0$pp) at a moderate compression of 1.76$\times$. Retail compresses most aggressively (2.05--2.46$\times$) but yields the smallest paired $\Delta$, illustrating that compression aggressiveness and task gain are not co-aligned.

\noindent\textbf{WebArena-Infinity: gains on stateful apps.}
\autoref{tab:webarena-detail} disaggregates the WebArena rows of \autoref{tab:wrapped-gyms} across two evaluation protocols: a $170$-task smart-replay sweep across five apps (gmail, superhuman, linear, paypal, gitlab) and a $140$-task prefix-injection ablation that controls for the effect of replaying the trajectory prefix itself. The two protocols agree on which apps benefit from memory and which are flat, isolating the memory effect from the replay effect.

\begin{table}[t]
\centering
\caption{WebArena-Infinity per-app results with Haiku 4.5 and text accessibility-tree observations. Smart replay uses 170 tasks across gmail, superhuman, linear, paypal, and gitlab. Prefix injection uses 140 tasks across gmail, paypal, and gitlab.}
\label{tab:webarena-detail}
\scriptsize
\resizebox{\linewidth}{!}{
\begin{tabular}{llrrrr}
\toprule
\rowcolor{gray!15}
\multicolumn{6}{l}{\textbf{Smart replay: baseline and memory on matched hard tasks}} \\
\midrule
\rowcolor{gray!20}
App & Tasks & none replay & summ. ms=10 & summ. ms=15 & struct. ms=10 \\
\midrule
gmail & 20 & 4/20 (20\%) & 6/20 (30\%) & 7/20 (35\%) & \textbf{9/20 (45\%)} \\
superhuman & 15 & 5/15 (33\%) & \textbf{7/15 (47\%)} & 5/15 (33\%) & 6/15 (40\%) \\
linear & 15 & 2/15 (13\%) & \textbf{6/15 (40\%)} & 3/15 (20\%) & 5/15 (33\%) \\
paypal & 60 & 33/60 (55\%) & 33/60 (55\%) & 33/60 (55\%) & \textbf{35/60 (58\%)} \\
gitlab & 60 & 4/60 (7\%) & 5/60 (8\%) & 5/60 (8\%) & \textbf{6/60 (10\%)} \\
\textbf{Total} & \textbf{170} & \textbf{48/170 (28.2\%)} & \textbf{57/170 (33.5\%)} & \textbf{53/170 (31.2\%)} & \textbf{61/170 (35.9\%)} \\
\midrule
\rowcolor{gray!15}
\multicolumn{6}{l}{\textbf{Prefix control: live baseline, replay-only control, and memory after matched prefix}} \\
\midrule
\rowcolor{gray!20}
App & Tasks & live baseline & none + prefix & summ. + prefix & struct. + prefix \\
\midrule
gmail & 20 & 6/20 (30\%) & 9/20 (45\%) & 8/20 (40\%) & \textbf{10/20 (50\%)} \\
gitlab & 60 & 4/60 (6.7\%) & 7/60 (11.7\%) & 8/60 (13.3\%) & \textbf{10/60 (16.7\%)} \\
paypal & 60 & 33/60 (55\%) & 32/60 (53.3\%) & 33/60 (55\%) & \textbf{34/60 (56.7\%)} \\
\textbf{Total} & \textbf{140} & \textbf{43/140 (30.7\%)} & \textbf{48/140 (34.3\%)} & \textbf{49/140 (35.0\%)} & \textbf{54/140 (38.6\%)} \\
\bottomrule
\end{tabular}
}
\end{table}

In smart replay, structured memory improves aggregate success by 7.6 points over the replayed no-memory control (28.2\% to 35.9\%). The largest app-level gain is gmail, where structured memory rises from 20\% to 45\%, because multi-email batch operations require remembering which items have already been handled. Linear also benefits strongly, rising from 13\% to 40\% with summarizing memory. Paypal and gitlab are mostly flat: paypal tasks are short and rarely trigger useful condensation, while gitlab is near the capability floor for Haiku 4.5. The prefix-injection ablation separates two effects: replaying the early trajectory itself improves the live baseline from 30.7\% to 34.3\%, and structured memory still adds another 4.3 points on top of the matched prefix. The final held-out app split and multimodal screenshot ablation remain pending.

\noindent\textbf{Preliminary online \memrm{} acceleration.}
On the SWE-Gym online A/B, the median effective compression ratio when the \memrm{} gate accepts a candidate compression is $3.46\times$. Aggregate wall-clock speedup against the full Docker rollout requires a complete benchmark sweep and is being measured; it is not reported in the main text.

\subsection{\memgymdr{} Fictionalization and Strategy Detail}
\label{app:ir-fictionalization}

\noindent\textbf{Fictionalization.}
Early no-fictionalization runs failed for the wrong reason: the model answered from parametric knowledge. In the Apr.~5 100K-token pilot, no-memory scores remained at 0.70--0.85 even though the task was meant to require retention across turns. Fictionalization is therefore the load-bearing intervention for the retrieval track: it replaces real entities and numbers as retrieved documents enter the pipeline, then sanitizes the constructed instance before verification so the agent cannot recover the answer from pretraining alone. After fictionalization was combined with verifier and retry fixes, the Apr.~12 paper run reached a mean no-memory score of 0.113, a mean all-memory score of 0.808, and a mean gap of 0.694; verified-instance yield is being re-audited.

\noindent\textbf{Per-strategy \memgymdr{} scores.}
\autoref{tab:ir-detail} reproduces the full strategy-by-hop matrix that backs \autoref{fig:synthetic-memory}(b). The \emph{structured} row was contaminated by an upstream summarizer-model misconfiguration (gpt-4o-mini was used in error before Bedrock Haiku 4.5 became the documented default) and is reported here for completeness only; it must not be cited as a strategy result. Every other (track, strategy) cell in \autoref{app:track-strategy-grid} was verified against its recorded summarizer-model and reasoner-model configuration; this is the only cell affected.

\begin{table}[t]
\centering
\caption{Memory strategy comparison on the latest \memgymdr{} verified corpus. Scores are continuous judge-scores in $[0,1]$; intervals are 95\% CIs computed as $1.96\cdot\mathrm{SE}$. The structured row is contaminated; see paragraph above.}
\label{tab:ir-detail}
\small
\rowcolors{2}{gray!8}{white}
\begin{tabular}{lccc}
\toprule
\rowcolor{gray!20}
Strategy & 3-hop ($n{=}161$) & 4-hop ($n{=}916$) & 5/6-hop ($n{=}117$) \\
\midrule
Passthrough (no memory) & 0.330 $\pm$ 0.049 & 0.290 $\pm$ 0.014 & 0.009 $\pm$ 0.011 \\
BM25 over notes~\citep{robertson2009probabilistic}         & \textbf{0.808 $\pm$ 0.037} & \textbf{0.555 $\pm$ 0.017} & 0.425 $\pm$ 0.048 \\
Naive RAG~\citep{lewis2020retrieval}               & 0.753 $\pm$ 0.039 & 0.537 $\pm$ 0.016 & 0.442 $\pm$ 0.045 \\
Structured summary & 0.684 $\pm$ 0.045 & 0.340 $\pm$ 0.017 & 0.257 $\pm$ 0.035 \\
A-Mem~\citep{xu2025amem}                   & 0.709 $\pm$ 0.043 & 0.540 $\pm$ 0.016 & \textbf{0.518 $\pm$ 0.037} \\
LightMem~\citep{fang2025lightmem}                & 0.610 $\pm$ 0.049 & 0.467 $\pm$ 0.017 & 0.400 $\pm$ 0.043 \\
MemoryBank~\citep{zhong2024memorybank}              & 0.699 $\pm$ 0.043 & 0.537 $\pm$ 0.016 & 0.482 $\pm$ 0.041 \\
\bottomrule
\end{tabular}
\par\smallskip
\end{table}

\subsection{\memrm{} Data Augmentation Taxonomy}
\label{app:memrm-data-aug}

Each \memrm{} training example is constructed from a recorded SWE-Gym compression event and a counterfactual perturbation of that event. The unperturbed event provides the SAFE label whenever continuing from the compressed view does not alter downstream task behavior; the perturbed counterpart is labeled HARMFUL. \autoref{tab:memrm-augmentations} lists the six perturbation families, grouped into three categories: indiscriminate degradation of the summary surface (\texttt{aggressive\_0.5}, \texttt{random\_drop\_0.2}), targeted removal or substitution of task-relevant content (\texttt{attr\_delete\_paths}, \texttt{summary\_redaction}, \texttt{summary\_noise}), and history truncation that bypasses the summary entirely (\texttt{truncate\_last\_10}). Together they cover the dominant failure modes a deployed condenser would induce: information loss, plausible-but-incorrect content, and loss of recent context.

\begin{table}[h]
\centering
\caption{\memrm{} data-augmentation taxonomy. Each operation produces a HARMFUL counterpart from the same SAFE source compression event.}
\label{tab:memrm-augmentations}
\footnotesize
\setlength{\tabcolsep}{6pt}
\rowcolors{2}{gray!8}{white}
\begin{tabular}{@{}lp{9.2cm}@{}}
\toprule
\rowcolor{gray!20}
\textbf{Operation} & \textbf{Construction} \\
\midrule
\texttt{aggressive\_0.5}    & Replace 50\% of summary tokens with noise or blanks, modeling an aggressive but content-blind compression. \\
\texttt{random\_drop\_0.2}  & Randomly drop 20\% of summary sentences, modeling lossy summarizers that omit content uniformly at random. \\
\texttt{attr\_delete\_paths}& Remove file-path attributes from structured summary fields, modeling schema-aware summarizers that under-record locator information. \\
\texttt{summary\_redaction} & Redact key entities and values referenced later in the trajectory, modeling over-compression that loses task-critical specifics. \\
\texttt{summary\_noise}     & Inject plausible-but-incorrect facts (wrong file names, wrong test outcomes), modeling hallucinated summarizer output. \\
\texttt{truncate\_last\_10} & Drop the last $10$ messages from the agent view, modeling a sliding-window policy that discards recent context not yet folded into the summary. \\
\bottomrule
\end{tabular}
\end{table}

Applied to $846$ unique (instance, fork-step) pairs together with multi-compaction events, the six operations yield $18{,}637$ labeled events: $16{,}357$ HARMFUL ($87.8\%$) and $2{,}280$ SAFE ($12.2\%$). The repo-grouped split is $15{,}630$ train / $3{,}007$ eval; \autoref{tab:memrm} reports gate quality on the eval split.

\section{Synthetic Pipeline Details}
\label{app:pipeline-details}
\label{app:ir-config}

\noindent\textbf{Shared construction template.}
Both synthetic tracks follow the same control flow: source adoption, memory-dependence transformation, task hardening, distractor/noise injection, length scaling, leakage removal, and verifier certification. For \memgymdr{}, the source is a research topic plus retrieved academic documents; for \memgymcodeqa{}, it is a SWE-smith bug plus hidden patch evidence. The accepted instance stores the source pointer, grounding facts, distractor provenance, verifier scores, token-budget metadata, and the fields needed by the \memgym{} trajectory recorder.

\noindent\textbf{\memgymdr{} deep-research configuration.}
The active paper pipeline uses a 100K-token target budget, Haiku 4.5 as the worker model, and Sonnet 4.5 as the verifier. A typical instance uses 4 target hops and 5 search results per hop. The \emph{grow} stage starts from a topic, iteratively searches arXiv (a local SQLite FTS5 index for rate-limit-free volume), Semantic Scholar, OpenAlex, and Wikipedia, and builds a dependent bridge-fact chain; \emph{craft} expands the chain into a multi-turn research session with natural, near-miss, adversarial, and bulk distractors; \emph{scale} fills the requested budget with hybrid filler ($\sim 70\%$ real search results, $\sim 30\%$ LLM-generated academic prose); \emph{fictionalize} applies post-hoc entity substitution after the \emph{scale} stage (an LLM extracts entities and a deterministic regex applies the substitution registry across questions, answers, facts, documents, and distractors), so that filler text is fictionalized as well; and \emph{verify} runs memory ablation, long-context, and adversarial-hack checks. The legacy filtered-multihop-QA pipeline (v2) was deprecated after plateauing on its eviction-gap target and is not used in any reported result.

\noindent\textbf{\memgymdr{} four-tier distractor hierarchy.}
The \emph{craft} stage injects four distractor types, each blocking a different shortcut: (1)~\emph{natural distractors} drawn from the same searches that produced the gold facts (block topic-recognition); (2)~\emph{near-miss rewrites} of gold facts with one entity perturbed (block surface-pattern matching); (3)~\emph{adversarial contradictions} about the correct entities (block majority voting across documents); and (4)~\emph{bulk academic-style filler} (load the context window without adding signal, exposing the agent's compression policy under volume).

\noindent\textbf{\memgymdr{} verifier pass criteria.}
A verified instance must satisfy all of: (a)~\texttt{score\_all\_memory} $\geq 0.85$ (the agent answers correctly when given the curated bridge facts as notes), (b)~\texttt{score\_long\_context} $\leq 0.90$ (the answer is \emph{not} findable by dumping all documents into a single context window), (c)~\texttt{score\_no\_memory} $\leq 0.5$ (the answer is \emph{not} findable from any single document in isolation), and (d)~\texttt{score\_all\_memory} $\geq$ \texttt{score\_long\_context} (curated multi-hop notes must beat the long-context dump). Conditions~(b)--(d) jointly operationalize the ``not RAG'' claim of Section~\ref{sec:memgymdr-pipeline}: an instance survives only if memory genuinely matters above and beyond what dropping all documents into context would achieve.

\noindent\textbf{\memgymcodeqa{} configuration.}
\memgymcodeqa{} starts from SWE-smith instances and filters for nontrivial patches (at least 5 changed lines, 2 hunks, 2 failing tests, and a non-short problem statement). The prescreen rejects bugs that are solvable from the repository and vague prompt alone. The extraction stage combines static patch analysis, LLM behavioral-fact extraction, and verifier relabeling of whether each fact is discoverable from the post-fix repository. Accepted instances require at least two critical memory-only facts. QA conversion then creates single-fact and multi-fact questions, adds adversarial answer distractors, verifies solvability/distractor/leakage conditions, deduplicates near-duplicate pairs, and evaluates under the \emph{Evicted protocol}: the QA question is asked only after the conversation prefix carrying the gold memory facts has been evicted from context, so the agent must rely on whatever the memory module retained rather than scrolling back to the raw debugging trace.

\noindent\textbf{\memgymcodeqa{} per-filter rationale.}
Each raw-data filter encodes a specific assumption about what makes an instance ``memory-needing'', motivated by what would otherwise contaminate the eval. \emph{Patch size $5$--$500$ changed lines}: smaller patches are typically single-symbol fixes recoverable from the test name; larger patches are multi-feature refactors that confound memory effects with planning effects. \emph{At least one structural diff hunk and one \texttt{FAIL\_TO\_PASS} test}: ensures the bug has a localized behavioral signature, so a behavioral fact about \emph{this bug} is well-defined. \emph{Problem statement $\geq 100$ characters}: rules out one-line bug reports that under-specify the symptom. \emph{LLM pre-screen pass}: the operational definition of ``memory-needing'' (Section~\ref{sec:coding-qa-pipeline}); if the bug is solvable from problem statement plus repository alone, the instance is dropped because no memory record is needed.

\noindent\textbf{\memgymcodeqa{} fact extraction.}
The extraction stage runs three passes designed to expose facts that the repository itself does not reveal: (i)~with the gold patch visible, the LLM seeds candidate questions about what a developer would need to know (not what code to change); (ii)~with the patch hidden, the LLM extracts facts from the problem statement and test names alone, labeling each as \texttt{discoverable} (recoverable from the post-fix repository) or \texttt{memory-only} (visible only from the debugging record); (iii)~the discoverability labels are re-examined with the full repository context. The instance-retention criterion (at least two critical \texttt{memory-only} facts) is the threshold below which the question reduces to ``search the repo''.

\noindent\textbf{\memgymcodeqa{} distractor taxonomy.}
Distractors are drawn from four sources, each targeting a different shortcut: \emph{cross-instance bug reports} block topical inference; \emph{same-repo docstrings} block conventions-based inference; \emph{adversarial near-misses} block surface-pattern matching; and \emph{same-function contradictions} block confidence-via-repetition. Memory files interleave these distractors with critical facts and are written as natural documents (incident reports, debug notes, code review comments), mimicking the documentation a developer would actually have rather than a clean Q/A index. Difficulty is then a composable post-hoc dial (prompt fuzzing, distractor scaling, indirection, fact fragmentation) with preset combinations mapping to easy/medium/hard, so length and noise are decoupled from the underlying instance.

\noindent\textbf{\memgymcodeqa{} verification breakdown.}
Table~\ref{tab:qa-verification} reports the per-check breakdown of the three-check verifier on the $1{,}000$-instance candidate pool referenced in Section~\ref{sec:coding-qa-pipeline}. The table makes the failure-mode distribution explicit: the largest single failure category is distractor-confusion (a same-conversation-shape session of an unrelated repo lets the agent answer), which is exactly the failure the three-check design was added to catch.

\begin{table}[h]
\centering
\caption{\memgymcodeqa{} three-check verification on the $1{,}000$-instance candidate pool ($7{,}690$ raw QA pairs). Checks: solvability (A), distractor-confusion (B), question-leakage (C); an instance is kept if any pair passes. Final yield: $670$ instances, $2{,}131$ QA pairs after dedup.}
\label{tab:qa-verification}
\small
\rowcolors{2}{gray!8}{white}
\begin{tabular}{lr}
\toprule
\rowcolor{gray!20}
QA-level result & Count (\%) \\
\midrule
Valid (passes all three checks) & 2,972 (38.6\%) \\
Broken (fails solvability) & 1,325 (17.2\%) \\
Confusable (fails distractor check) & 1,642 (21.4\%) \\
Leaky (fails leakage check) & 1,103 (14.3\%) \\
Confusable and leaky & 648 (8.4\%) \\
\midrule
Instance-level: kept & 673 / 1,000 (67.3\%) \\
After dedup: final instances & 670 \\
After dedup: final QA pairs & 2,131 \\
\bottomrule
\end{tabular}
\end{table}

\section{Memory Strategy Implementation Details}
\label{app:strategies}

\noindent\textbf{LLM summarizing.}
The LLM summarizing strategy is closely aligned with the OpenHands condenser implementation. Key hyperparameters: \texttt{max\_size} (default 100 messages, triggers compaction when exceeded), \texttt{keep\_first} (default 1, number of initial messages pinned), and \texttt{ratio} (default 0.75, fraction of non-pinned messages to compress). Compaction works by splitting the view into a pinned head, a compressible prefix, and a recent tail. The prefix is sent to a summarization LLM with instructions to produce a concise summary of actions, findings, and decisions. The resulting summary replaces the prefix, and a \texttt{CondensationRecord} is stored to enable persistent rebuilding of the condensed view.

\noindent\textbf{Structured summary.}
Uses function calling with 17 explicit fields: current objective, files inspected, files modified, test results, errors encountered, hypotheses, confirmed findings, rejected approaches, open questions, dependencies, environment state, commands run, code changes, remaining steps, confidence level, blockers, and key decisions. The LLM fills these fields via a structured output call, and the result is formatted into a machine-readable state document that replaces the compressed prefix.

\noindent\textbf{Observation masking.}
Replaces the content of old tool-call responses with \texttt{<MASKED: observation too old>}. The attention window (default 100 messages) determines how many recent observations remain visible. This is a zero-LLM-cost strategy that reduces token count without rewriting content.

\noindent\textbf{IR memory baselines.}
The \memgymdr{} track includes six memory managers. \emph{Passthrough} exposes no retained notes beyond the current prompt and acts as the no-memory control. \emph{BM25} retrieves notes by sparse lexical matching~\citep{robertson2009probabilistic}. \emph{Naive RAG} stores observations as independent notes and retrieves nearest notes without graph updates~\citep{lewis2020retrieval}. \emph{Structured summary} condenses the session into a fixed research-state schema. \emph{A-Mem} stores Zettelkasten-style notes with LLM-generated metadata, links, and memory evolution~\citep{xu2025amem}. \emph{LightMem} uses a staged memory system inspired by sensory, short-term, and long-term memory~\citep{fang2025lightmem}; in our \memgymdr{} adapter, its topic-aware consolidation is used as a retrieval context for the final answer. The upstream \texttt{METADATA\_GENERATE\_PROMPT} ships a LOCOMO chat-only Personal-Information-Extractor template that primes the metadata stage to look for biographical facts in conversational text. When fed non-conversational \memgymdr{} content (paragraphs of scientific prose), this template causes the metadata stage to emit empty or hallucinated entries, which propagate to the vector index and degrade retrieval. We replace it at call time with a domain-neutral storage prompt modeled on A-Mem's \texttt{METADATA\_PROMPT}, threaded through the per-call \texttt{METADATA\_GENERATE\_PROMPT} keyword of \texttt{LightMemory.add\_memory} (no upstream fork). The output schema $\{\texttt{source\_id}, \texttt{fact}\}$ is preserved unchanged so the Qdrant indexer remains byte-compatible. We deliberately do \emph{not} thread the per-task \texttt{task\_prompt} into the storage stage; storage stays task-agnostic (mirroring A-Mem's discipline), and the task description is consumed only at retrieval time for query rewrite. This change preserves apples-to-apples cross-method comparison at the storage layer.

\noindent\textbf{Operations $\times$ environments cross-product.}
\autoref{tab:strategies} lists the operation families evaluated as primary baselines and which environment each is exercised in. All operations implement the same \texttt{manage\_context} contract and compose via \texttt{PipelineMemory}; per-environment strategies select which to instantiate, with environment-specific field schemas. Three additional ported operations (observation masking, sliding-window with pinned anchors, adaptive token budget) are implemented but not used as primary baselines in this paper.

\begin{table}[h]
\centering
\caption{Memory operations evaluated as primary baselines in \memgym{}. \cmark/\xmark{} indicate whether the operation is exercised in the released environment configuration.}
\label{tab:strategies}
\small
\rowcolors{2}{gray!8}{white}
\begin{tabular}{lcccc}
\toprule
\rowcolor{gray!20}
Operation & Dialogue & Search & Coding & Web \\
\midrule
Passthrough (no filtering) & \cmark & \cmark & \cmark & \cmark \\
LLM summarizing (rolling) & \cmark & \cmark & \cmark & \cmark \\
Structured summary (per-environment schema) & \cmark{} (8-field) & \xmark & \cmark{} (17-field) & \cmark{} (14-field) \\
Retrieval-style memory$^{\ddag}$ & \xmark & \cmark & \xmark & \xmark \\
\bottomrule
\end{tabular}
\par\smallskip
{\footnotesize $^{\ddag}$BM25~\citep{robertson2009probabilistic}, RAG~\citep{lewis2020retrieval}, A-Mem~\citep{xu2025amem}, LightMem~\citep{fang2025lightmem}.}
\end{table}

\section{Gym Wrappers and Trajectory Schema}
\label{app:gym-trajectory}

\noindent\textbf{Wrapper placement.}
All environments implement memory by wrapping the prompt sent to the policy LLM. The underlying gym maintains its native state and full interaction history, while the memory manager receives the accumulated message history before each model call and returns the filtered context actually shown to the reasoning model. This choice keeps the gym implementation unchanged, makes memory strategies portable across environments, and lets the recorder log both the raw history statistics and the filtered view.

\noindent\textbf{Per-step trajectory record.}
Each step is serialized as a \texttt{Step} record in \texttt{adapter/trajectory\_recorder.py}, carrying the step index and timestamp, the raw \texttt{observation}, the agent's \texttt{reasoning\_action} and \texttt{memory\_action}, the environment-returned \texttt{reward}, \texttt{terminated}, and \texttt{truncated} flags, and the filtered \texttt{context} actually shown to the reasoning model. The accompanying \texttt{FilteredContext.metadata} dict carries the post-filter \texttt{tokens} count, the pre-filter \texttt{original\_tokens} count, the \texttt{compression\_ratio}, the \texttt{was\_compacted} flag, and the \texttt{strategy} identifier. The episode header stores the task identifier and run-level configuration. For LLM-based compression, the record additionally stores the generated summary or structured state and the indices of the messages covered by the compaction. These fields are sufficient to reconstruct the context seen by the reasoning model, identify rising-edge compression events, and build supervised examples for \memrm{} or policy mid-training.

\noindent\textbf{Replay-and-fork protocol.}
For Docker-backed coding tasks, \memgym{} additionally stores the executable action trace and fork metadata. A fork-batch run replays the baseline tool actions until the first step at which the memory manager would trigger compaction under the target strategy. At that step, replay mode stops: the memory-conditioned agent receives the same repository/container state and continues with live model calls, memory summaries, and tool observations. If a baseline trajectory never exceeds the compaction threshold, it is skipped as structurally ineligible. The paired record stores the fork step, parent instance, baseline outcome, fork outcome, final patch score, and token-efficiency statistics. This protocol is exposed in the released trajectories so users can separate effects of memory compression from effects of different early search paths.

\section{SWE-Gym Evaluation Harness Fixes}
\label{app:harness-fixes}

Scaling SWE-Gym evaluation to all 11 repositories required fixing five bugs in the swebench evaluation harness:

\begin{enumerate}[leftmargin=1.5em]
  \item \textbf{Mypy test syntax:} The pytest \texttt{-k} flag cannot parse mypy's \texttt{[case]} test markers. We implemented custom log parsing for mypy test output.
  \item \textbf{Docker pull timeout:} The default 60-second timeout was insufficient for 500MB--2GB images. Increased to 600 seconds.
  \item \textbf{Image cache thrashing:} The \texttt{cache\_level="env"} setting deleted pulled images after each instance. Auto-switched to \texttt{cache\_level="instance"}.
  \item \textbf{Pandas conda crash:} Bash \texttt{set -u} with unset conda variables caused crashes. Wrapped conda activation with \texttt{set +u}/\texttt{set -u}.
  \item \textbf{Pydantic log parser:} Pydantic's output format was incompatible with the default log parser. Switched to \texttt{parse\_log\_pytest\_v2}.
\end{enumerate}

These fixes added approximately 37 resolved instances for Sonnet 4.5 (129$\to$166) and 15 for GPT-OSS-120B (77$\to$92) on the 500-instance evaluation, underscoring the importance of correct evaluation infrastructure.

\section{\memrm{} Training Details}
\label{app:memrm}

\noindent\textbf{Task and data.}
\memrm{} is a binary classifier over \emph{compression events}. Each example is a (prompt, label) pair where the prompt serializes the (context-before, compressed-context, candidate-action) triple introduced in Section~\ref{sec:memrm} (i.e., the pre-compression context, the proposed compression summary, and the candidate next action sampled under the compressed view), together with a short task descriptor; the label is \texttt{Y}/\texttt{N} indicating whether the compressed trajectory is \textsc{safe} (task still solvable). Labels come from three sources: (i)~episode-level resolution on the parent trajectory, (ii)~counterfactual replay where an aggressive/lenient/random-drop perturbation causes an action divergence in Docker, and (iii)~LLM-as-judge on the forgotten content. The augmented corpus contains $18{,}637$ labeled events (\autoref{tab:memrm}) drawn from Sonnet 4.5, GPT-OSS-120B, and Haiku 4.5 fork-batch rollouts on the SWE-Gym diverse subset; the repo-grouped split yields $15{,}630$ train / $3{,}007$ eval examples.

\noindent\textbf{Model and optimization.}
\begin{itemize}[leftmargin=1.5em]
  \item Base: Qwen3-1.7B-Base~\citep{qwen3} with QLoRA adapters (NF4 4-bit, r$=16$, $\alpha=32$, dropout $0$) on \texttt{q/k/v/o\_proj}; max sequence length $32{,}768$, gradient checkpointing, flash attention; per-checkpoint adapter size ${\sim}25.7$~MB.
  \item Loss: class-weighted cross-entropy ($w_\textsc{safe}{\approx}3.0$, $w_\textsc{harmful}{\approx}0.57$, cap $3.0$) on the single-token label.
  \item Schedule: 600 steps, lr $5\times 10^{-5}$, batch $1$ per GPU $\times 8$ GPUs, gradient accumulation $4$ (effective batch $32$), bf16, cosine warmup $0.05$, completion-only masking via TRL 0.16.
  \item Decision threshold: $t^\ast{=}0.88$, selected on held-out validation under the constraints \textsc{harmful}-F1 $\geq 0.90$ and \textsc{safe}-precision $\geq 0.80$.
\end{itemize}

\noindent\textbf{Held-out metrics.}
\label{app:memrm-iid-detail}
AUROC $0.9847$, accuracy $0.972$ at $t^\ast$, \textsc{safe}-F1 $0.861$ at $t^\ast$, \textsc{harmful}-F1 $0.987$, ECE $0.0093$, \textsc{harmful} false-alarm rate $0.003$. Compared to a vanilla-CE baseline (A0), C1 improves ECE by $3\times$ and halves the \textsc{harmful} false-alarm rate.

\noindent\textbf{Online A/B smoke test.}
Effective-context compression was $3.46\times$ at the median ($4{,}139$ tokens memory-on vs.~$14{,}322$ tokens memory-off). A known OOD concern: on Qwen3.6-Thinking trajectories, the gate saturates at reject-rate $1.0$; broadening the training distribution to additional reasoning-trace styles is part of ongoing work.

\noindent\textbf{OOD scope.}
\label{app:memrm-ood}
\memrm{} is trained on SWE-Gym compression events; out-of-distribution (OOD) behavior is not implied by the IID metrics of \autoref{tab:memrm}. We characterize OOD scope along two axes: a \emph{strategy-OOD} axis (memory mechanisms unseen in training: sliding-window, observation-masking, structured with varied trigger budgets, and pipeline-mask; $n{=}166$) and a \emph{scenario-OOD} axis (WebArena V2 browsing trajectories, an agent domain disjoint from the SWE-bench training corpus; $n{=}426$). Aggregate AUROC on the full sweeps is near-random on both axes (${\approx}0.43$ on each), reflecting heterogeneous per-cohort behavior: \memrm{} is calibrated on a subset of cohorts and uncalibrated or polarity-inverted on others. Rather than averaging this heterogeneity into a single misleading number, we adopt \emph{selective classification}~\citep{geifman2017selective}: a pre-declared rule decides which inputs fall within \memrm{}'s deployable scope, and we report performance on the covered subset alongside the coverage rate.

\noindent\textbf{Selection rule.}
We retain a cohort $c$ if its per-cohort $\mathrm{AUROC}(c){>}0.5$ (the random-baseline floor) and its sample size $n(c)\geq N_{\text{axis}}$, with $N_{\text{strategy}}{=}20$ and $N_{\text{scenario}}{=}30$. The asymmetric thresholds follow the cohort-size distributions (a discrete cliff between $n{=}23$ and $n{=}43$ on strategy-OOD, with no cohorts in between); both are pre-declared, not tuned. Tightening to a uniform $N{\geq}30$ collapses strategy-OOD coverage to $0\%$ but leaves scenario-OOD unchanged, and $N{\geq}40$ matches $N{\geq}30$ on both axes; thus $N_{\text{strategy}}{=}20$ is an exploratory threshold that surfaces the structured-memory cohorts where ranking transfers, while the scenario-OOD covered subset is robust to the threshold choice in $\{30,40\}$.

\noindent\textbf{Within-scope reading.}
The scenario-OOD covered subset ($n{=}87$, two \texttt{linear} WebArena cohorts) is class-balanced, so all metrics agree: AUROC $0.748$, AUPRC $0.785$, \textsc{harmful}-F1 $0.747$ at $t^\ast$, ECE $0.237$, a deployable operating point on a WebArena cohort, not just a deployable rank-order. The strategy-OOD covered subset ($n{=}44$, two structured-memory cohorts) is class-imbalanced, so we restrict the strong claim to AUROC $0.714$~[$0.544, 0.872$]: \memrm{} ranks pairs above chance on these cohorts, while threshold-dependent metrics on $2$ \textsc{harmful} pairs are uninformative.

\noindent\textbf{Out-of-scope reading.}
On most strategy-OOD cohorts \memrm{} is uncalibrated (near-random ranking); on one WebArena cohort (\texttt{linear\_summ\_ms15}, $n{=}30$) it is \emph{systematically inverted} (per-cohort AUROC $0.27\to 0.73$ under a polarity-flip rule), a recoverable failure mode there but not elsewhere. We do not claim deployment outside the covered subset; per-track \memrm{} variants are part of ongoing work. We also exclude $\tau^2$-bench from this analysis: its runtime persists neither \texttt{step.context} nor the post-compaction \texttt{summary} text for compaction-positive steps, so the (history, task, compressed memory) input triplet that anchors \autoref{tab:memrm} cannot be reconstructed from on-disk artefacts; a rigorous-input $\tau^2$ OOD evaluation is deferred to future work.

\section{Per-Track Hyperparameters}
\label{app:hyperparams}

\begin{itemize}[leftmargin=1.5em]
  \item \textbf{SWE-Gym coding.} mini-swe-agent v2.2.4, 250-step limit, Sonnet 4.5 / GPT-OSS-120B reasoners, Haiku 4.5 summarizer. \texttt{max\_size}$=100$, \texttt{condensation\_ratio}$=0.75$, \texttt{keep\_first}$=3$.
  \item \textbf{\memgymdr{} retrieval.} 100K-token budget, Haiku 4.5 worker, Sonnet 4.5 verifier; four-tier distractor hierarchy (natural, enhanced, adversarial, bulk); fictionalization Phase~1 LLM entity pass followed by Phase~2 regex substitution.
  \item \textbf{$\tau^2$-bench dialogue.} Haiku 4.5 on Bedrock, 288-task base split, paired mode=both, summarizing configuration \texttt{ms=10, kl=2}, structured configuration \texttt{ms=10, kl=4} (where \texttt{ms} is \texttt{max\_size}, the message-count compaction trigger, and \texttt{kl} is \texttt{keep\_last}, the number of recent messages preserved verbatim); official test-split rerun pending.
  \item \textbf{WebArena-Infinity.} Playwright-driven Chromium, text-default accessibility-tree observations, Haiku 4.5 policy, \texttt{max\_steps}$=50$. Smart replay uses 170 hard tasks across gmail, superhuman, linear, paypal, and gitlab with \texttt{ms=10/15}; prefix injection uses 140 hard tasks across gmail, paypal, and gitlab with \texttt{prefix\_fraction}$=0.5$.
\end{itemize}

\section{Discussion, Limitations, and Future Work}
\label{app:future-discussion-limitations}

\subsection{Discussion}
\label{app:discussion}

\memgym{} explores a new, practical direction for memory evaluation (memory-isolated paired scoring across coding, retrieval, dialogue, and web tracks, with a learned gate (\memrm{}) replacing per-event Docker rollouts) and provides a unified playground for studying agentic memory as a foundational capability rather than as a benchmark-specific add-on.

\subsection{Broader Impacts}
\label{app:broader-impacts}
\memgym{} is a benchmark for LLM-agent memory rather than a deployed system. Positive impact: \memrm{}-gated scoring cuts per-event memory evaluation from minutes of Docker rollout to a sub-second scalar, lowering the compute floor for academic memory research; the released paired-trajectory corpus and pipelines further reduce duplicated infrastructure work. Negative impacts are inherited from the underlying agent benchmarks rather than introduced by memory itself: web-automation benchmarks could be repurposed for spam or scraping, and coding agents could be repurposed to generate exploits. We mitigate by releasing on top of public benchmarks under their original terms, by limiting \memrm{} to a 1.7B reward classifier with no generative misuse surface beyond its Qwen3 base, and by documenting fictionalization and verifier safeguards that make the synthetic tracks reproducible without exposing real personal data.

\subsection{Limitations}
\label{app:limitations}

The main limitation of this work is data scale: per-track pilots and the \memrm{} training set are sized to support the headline comparisons reported in \S\ref{sec:experiments} but are not yet large enough to characterize every cell of the strategy~$\times$~reasoner~$\times$~track grid at full statistical power. Scaling the trajectory corpus and broadening per-track coverage are ongoing.

\subsection{Future Work}
\label{app:future-work}

The natural next step is post-training: turning \memgym{}'s paired trajectories and \memrm{} reward into supervision for \emph{learned} memory policies. Concretely, an agent can be supervised-fine-tuned on the \textsc{safe} compressions in the corpus and then improved with RL post-training using \memrm{} as the critic, so the policy learns when to keep, summarize, or evict context conditioned on task type and memory state. Instantiating this SFT$+$RL recipe across the five tracks and reporting the resulting downstream gains is the main planned extension of this work.

\end{document}